\newif\ifconfver
\confverfalse      

\newif\ifcutshort      
\cutshorttrue

\newif\ifcutshortlvltwo  
\cutshortlvltwofalse

\ifconfver
\documentclass[10pt,twocolumn,twoside]{IEEEtran}
\else
\documentclass[11pt, draftclsnofoot, onecolumn]{IEEEtran}
\fi

\usepackage{amsmath,amsfonts,bm,amssymb,epsfig,psfrag,ifthen,color}
\usepackage{booktabs}
\usepackage{multirow}
\usepackage[table]{xcolor}  
\usepackage{array}
\usepackage{pifont}
\usepackage{calc}
\usepackage{caption}
\usepackage{graphicx}
\usepackage{psfrag}
\usepackage[tight,footnotesize]{subfigure} 
\usepackage{stfloats}
\usepackage{url}
\usepackage{longtable}
\usepackage{amsmath,amsfonts,amssymb,amsbsy,bm,paralist,theorem,ifthen,color}
\usepackage{algorithmic,algorithm,theorem}
\usepackage{mathtools}

\newcommand{\cmark}{\ding{51}}%
\newcommand{\xmark}{\ding{55}}%

\newcommand\Yc{\ensuremath{\mathcal{Y}}}
\newcommand\Dc{\ensuremath{\mathcal{D}}}

\newcommand\Lc{\ensuremath{{\mathcal{L}}}}
\newcommand\Pc{\ensuremath{{\mathcal{P}}}}
\newcommand\Ac{\ensuremath{{\mathcal{A}}}}

\newcommand\Oc{\ensuremath{{\mathcal{O}}}}

\newcommand\xb{\ensuremath{{\bf x}}}

\newcommand\yb{\ensuremath{{\bf y}}}

\newcommand\ub{\ensuremath{{\bf u}}}

\newcommand\bb{\ensuremath{{\bf b}}}

\newcommand\vb{\ensuremath{{\bf v}}}

\newcommand\zb{\ensuremath{{\bf z}}}

\newcommand\lambdab{\ensuremath{{\bm \lambda}}}

\newcommand\zerob{\ensuremath{{\bm 0}}}

\newcommand\E{\ensuremath{{\mathbb{E}}}}

\newcommand\Prob{\ensuremath{{\rm Prob}}}

\newcommand\Rbb{\ensuremath{{\mathbb{R}}}}
\newcommand\st{\ensuremath{{\rm s.t.}}}
\newcommand\Xc{\ensuremath{{\mathcal{X}}}}
\newcommand{\wt}{\widetilde}

\newcommand{\ol}{\overline}
\newcommand{\ul}{\underline}

\newtheorem{Lemma}{Lemma}

\newtheorem{Theorem}{Theorem}

\newtheorem{Corollary}{Corollary}

\newtheorem{Rmk}{Remark}
\newtheorem{assumption}{Assumption}

\newcommand{\tabincell}[2]{\begin{tabular}{@{}#1@{}}#2\end{tabular}}

\ifconfver

\else

\fi

\begin{document}

    \bibliographystyle{IEEEtran}

    \title{ {Beyond ADMM: A Unified Client-variance-reduced Adaptive Federated Learning Framework}}

    \ifconfver \else {\linespread{1.1} \rm \fi

        \author{\vspace{0.8cm}Shuai~Wang,
        	~Yanqing Xu, ~Zhiguo Wang, ~Tsung-Hui Chang, \\~Tony Q. S. Quek,
        	~and Defeng Sun\\
        	\thanks{Shuai Wang and Tony Q. S. Quek are with the Pillar of Information Systems Technology and Design, Singapore University of Technology and Design, 487372 Singapore (e-mail: shuaiwang@link.cuhk.edu.cn, tonyquek@sutd.edu.sg).}
            \thanks{Yanqing Xu and Tsung-Hui Chang (Corresponding author) are with the School of Science and Engineering, The Chinese University of Hong Kong, Shenzhen 518172, China (e-mail: xuyanqing@cuhk.edu.cn,~changtsunghui@cuhk.edu.cn).}
 \thanks{Zhiguo Wang is with College of Mathematics, Sichuan University, Chengdu, Sichuan 610064, China (e-mail: wangzhiguo@scu.edu.cn).}
 \thanks{Defeng Sun is with the Department of Applied Mathematics, The Hong Kong Polytechnic University, Hong Kong (e-mail: defeng.sun@polyu.edu.hk).}
            }

        \maketitle
        
\vspace{-1.5cm}
\begin{center}
\today
\end{center}\vspace{0.5cm}

        \begin{abstract}
	       As a novel distributed learning paradigm, federated learning (FL) faces serious challenges in dealing with massive clients with heterogeneous data distribution and computation and communication resources. Various client-variance-reduction schemes and client sampling strategies have been respectively introduced to improve the robustness of FL. Among others, primal-dual algorithms such as the alternating direction of method multipliers (ADMM) have been found being resilient to data distribution and outperform most of the primal-only FL algorithms. However, the reason behind remains a mystery still. In this paper, we firstly reveal the fact that the federated ADMM is essentially a client-variance-reduced algorithm. While this explains the inherent robustness of federated ADMM, the vanilla version of it lacks the ability to be adaptive to the degree of client heterogeneity. Besides,  the global model at the server under client sampling is biased which slows down the practical convergence. To go beyond ADMM, we propose a novel primal-dual FL algorithm, termed FedVRA, that allows one to adaptively control the variance-reduction level and biasness of the global model. In addition, FedVRA unifies several representative FL algorithms in the sense that they are either special instances of FedVRA or are close to it. Extensions of FedVRA to semi/un-supervised learning are also presented. Experiments based on (semi-)supervised image classification tasks demonstrate superiority of FedVRA over the existing schemes in learning scenarios with massive heterogeneous clients and client sampling.
            \\\\
        \end{abstract}


        \ifconfver \else \IEEEpeerreviewmaketitle} \fi

    \vspace{-0.5cm}

\section{Introduction}

As a local-privacy-aware distributed paradigm, federated learning (FL) recently has drawn significant attention in the distributed machine learning (ML) community \cite{FL_Beyond_2015}. In a typical FL setting,  a central server coordinates $N$ distributed clients to jointly solve the following learning problem:
\begin{align}\label{eqn: vanilla FL}
	\min_{\substack{\xb }}~&  f(\xb) \triangleq \sum_{i = 1}^{N} \omega_i f_i(\xb),
\end{align}
\noindent where $f_i(\xb)$ is the (possibly non-convex) local cost 
function and $\omega_i$ is the weight coefficient associated with client $i$.  
Different from traditional distributed learning \cite{ChangSPM2020}, FL faces numerous challenges such as limited communication resources,  and data heterogeneity and system heterogeneity over massive clients \cite{FML_CA_2019, FL_Ondevice_2016}.
Many FL algorithms including the popular FedAvg \cite{FedAvg_noniid_2019, CE_DDNN_2017} adopt the partial client participation ({PCP}) strategy using client sampling and local stochastic gradient descent ({local SGD}) \cite{LocalSGD_2019} to overcome the network congestion problem and improve the communication efficiency.
However, FedAvg is inefficient in dealing with data heterogeneity and system heterogeneity where the clients have non-i.i.d.  local datasets and have varied computational capabilities, respectively \cite{FedProx_2018}.  On one hand,  the presence of non-i.i.d data would cause the client drift issue which degrades the algorithm convergence performance and can even lead to model divergence \cite{FedAvg_noniid_2019,  SCAFFOLD_2020}.   
On the other hand, heterogeneity in the computational speeds results in large variations in the number of local updates performed by each client (i.e.,  heterogeneous local updates (HLU)), which leads to solution bias and convergence slowdown \cite{FedNova_2020}. In addition,  under PCP,  many of the FL algorithms require unbiased client sampling, otherwise the global model at the server is biased and the algorithm cannot converge to a proper solution \cite{Fed_PowerOfChoiceSelection}. 

To address the data heterogeneity issue, client-variance-reduction (CVR) schemes such as VRL-SGD \cite{VRL_SGD_2020} and SCAFFOLD \cite{SCAFFOLD_2020} have been proposed.
They identified the inter-client variance \footnote{If the local data of all clients have an identical distribution, then all local stochastic gradients (SGs) have the same mean and it is equal to the global SG. However, when local data have different distributions, there is a gap between the local SGs and global SG, which is called inter-client variance. Such a notion is widely used in the FL literature \cite{FedNova_2020}.} caused by non-i.i.d data as the main reason for the client drift issue and attempted to alleviate it. Specifically, in the CVR schemes, control variates are introduced to perturb the local stochastic gradient (SG) so as to approximate the \emph{global SG} (which assumes data samples are drawn from all client's data in the centralized manner),  thereby mitigating the client drift effect. For overcoming the system heterogeneity,  FedNova \cite{FedNova_2020} proposed the use of normalized averaging SGD to account for HLU. However,  the aforementioned methods do not fully resolve both the data and system heterogeneity issues.  For example,  VRL-SGD and SCAFFOLD do not consider HLU to address the system heterogeneity,  and FedProx and FedNova still suffer from convergence slowdown caused by non-i.i.d data. 

Interestingly,  recent findings show that primal-dual FL methods based on the alternating direction method of multipliers (ADMM)  \cite{ADMM_2010} \cite{NESTT_2016} are inherently resilient to both data and system heterogeneity,  see,  e.g.,  FedPD \cite{FedPD_2021}, FedADMM \cite{FedADMM_2022} and FedDyn \cite{FedDyn_2021}.   However,  their convergence rely on the constant and uniform client sampling,  and the requirement of the clients to either solve the local subproblems globally or to a sufficient accuracy.  Besides, it is not clear how the distributed ADMM algorithms are related to the existing CVR schemes.  Table \ref{table: com_FL_alg} summarizes the aforementioned FL algorithms with their client sampling strategies under PCP,  convergence rate,  and whether the clients adopt local SGD (LSGD) and HLU.
\subsection{Contributions}

In this paper, we are interested in developing FL algorithms that are not only robust against the aforementioned client heterogeneity issues,  but also are "adaptive" to the level of them. 
To this end, we firstly show that the federated ADMM algorithm is in fact a client-variance-reduction scheme, which explains its superior resilience when compared to the primal-only FL algorithms. 
Inspired by this, we propose a novel primal-dual FL algorithm, termed Federated Variance-Reduction Adaptive (FedVRA), that not only enjoys all the advantages of federated ADMM, but also is adaptive to the degree of client heterogeneity and client sampling schemes.
This is achieved by introducing two novel stepsize parameters in the dual variable update and the global variable update,  which respectively controls the client-variance-reduction level and biasness of the global model.
Convergence analysis shows that FedVRA can converge to a stationary solution in a sublinear rate under (time-varying) HLU and an arbitrary client sampling scheme,  as shown in Table \ref{table: com_FL_alg}.
Even intriguingly,  we show that FedVRA has an intimate relation with the aforementioned FL algorithms in the sense that they are either special instances of FedVRA or close to it. Lastly,  we extend FedVRA to solve a class of semi/un-supervised problems which involve two blocks of variables and variable constraints. 
Extensive experiments based on a (semi-)supervised image classification task demonstrate the superiority of FedVRA over the existing schemes.

\begin{table}[t!] 
	\centering \normalsize
	\caption{Comparison of representative FL algorithms. 
	}
	\setlength{\tabcolsep}{4.5mm}
	\label{table: com_FL_alg}
	\begin{tabular}{|c|c|c|c|c|c|}
		\hline    \rowcolor{gray!50}       
		Algorithm      &      PCP      & LSGD   &   HLU   &   Convergence rate \\  \hline\hline 
		FedAvg     &   unbiased &  \cmark    &  \xmark &    \Oc\bigg($\frac{\frac{\sigma^2}{S} + G^2}{\sqrt{mR}} + \frac{G}{R^{2/3}} + \frac{B^2}{R}$\bigg)  \\ \cline{1-5}
		FedProx         &     unbiased       &  \xmark &    \cmark  &$\mathcal{O}(\frac{B^2}{R})$ \\ \cline{1-5}
		VRL-SGD       &   \xmark     &\cmark    &    \xmark   &     \Oc$\bigg(\frac{\sigma^2}{S\sqrt{NR}}+ \frac{N}{R} \bigg)$ \\ \cline{1-5}
		FedNova      &     unbiased      & \cmark &  \cmark   & \Oc\bigg($\frac{\frac{\sigma^2}{S} + G^2}{\sqrt{mR}} + \frac{m(\frac{\sigma^2}{S} + G^2)}{R}$\bigg)\\ \cline{1-5}
		SCAFFOLD       &    unbiased      &\cmark   & \xmark &  \Oc\bigg($\frac{\sigma^2}{S\sqrt{mR}}+\frac{1}{R}\big(\frac{N}{m}\big)^{\frac{2}{3}}$\bigg)\\ \cline{1-5}
		FedDyn  & unbiased   & \xmark & \xmark  & $\mathcal{O}(\frac{N}{mR})$ \\ \cline{1-5}	
		FedPD  &\xmark  & \xmark & \cmark & $\mathcal{O}\bigg(\frac{\sigma^2 }{S} +\frac{1}{R} +  \epsilon\bigg)$ \\ \cline{1-5}	
		\tabincell{c}{Our paper   \\  FedVRA} &  \tabincell{c}{arbitrary   \& TV}  & \cmark & \tabincell{c}{\cmark   \& TV}   & $\mathcal{O}\bigg(  \frac{N\sigma^2}{mS} + \frac{N}{mR} \bigg)$ \\    
		\hline
	\end{tabular}

	$R$: Num. of rounds,  $m$: Num. of clients sampled, $S$: mini-batch size, $(G, B)$ bounds gradient dissimilarity \cite{SCAFFOLD_2020},  $\epsilon$: solution accuracy of local subproblems,  TV: time-varying
\end{table}

\subsection{Notations and assumptions}\label{sec: problem formulation}

Before proceeding, we summarize the key notations in Table \ref{table0001}, and make the following standard assumptions.
\begin{assumption}[\textbf{Lower boundedness and L-smoothness}] \label{assumption: lower-bounded}
	Each local cost function $f_i(\cdot)$ in problem \eqref{eqn: vanilla FL} is lower bounded, i.e., $f_i(\xb) \geq \underline{f} > {-\infty}$ and $L_i$-smooth, which implies $\|\nabla f_i(\xb) - \nabla f_i(\xb^\prime)\| \leq L_i\|\xb - \xb^\prime\|, \forall \xb, \xb^\prime$. 
\end{assumption}
\begin{assumption}[Bounded SGD variance] \label{assumption: SGD_variance}
	For a data sample $\xi_i$ uniformly sampled at random from $\Dc_i$, the resulting stochastic gradients (SG) for problem \eqref{eqn: vanilla FL} is unbiased and have bounded variances, i.e.,
	\begin{align}
		& \E[\nabla f_i(\xb_i; \xi_i)] = \nabla f_i(\xb_i), \\
		&\E[\|\nabla f_i(\xb_i; \xi_i) - \nabla f_i(\xb_i)\|^2] \leq \sigma^2, 
	\end{align}where $\sigma > 0$ is a constant.
\end{assumption}
Let  $g_i(\xb_i)$ denote the SG of $f_i(\cdot)$ at $\xb_i$ over a mini-batch of $S$ samples i.i.d drawn from the client $i$'s dataset, i.e., $g_i(\xb_i) \triangleq \frac{1}{S} \sum_{\xi_i} \nabla f_i(\xb_i; \xi_i)$. Then, $g_i(\xb_i)$ is unbiased with variance bounded by $\frac{\sigma^2}{S}$.

\begin{table}[t!] 
	\caption{Summary of Notations}\label{table0001}
	\centering \normalsize
	\setlength{\tabcolsep}{7mm}
	\begin{tabular}{ll}\hline\hline
		{\bf Notation} & {\bf Definition} \\ \hline
		$\Ac^r$ ($|\Ac^r| = m$) & Subset of clients sampled in round $r$ \\
		$p_i^r$ & Probability of client $i$ being sampled in round $r$ \\
		$\xb_i^{r, t}$ & Variable $\xb_i$ of client $i$ at iteration $t$,  round $r$ \\
		$\xb_i^{r+1}$ & Updated variable $\xb_i$ of client $i$ in round $r$ \\
		$\xb_0^r$  & Global variable $\xb_0$ at the server in round $r$\\
		$Q_i^r$ & Number of local updates w.r.t. $\xb_i$ in round $r$ \\
		$\nabla f_i(\xb_i; \xi_i)$ & SG of $f_i(\cdot)$ w.r.t $\xb_i$ and a sample $\xi_i \in \Dc_i$ \\
		$g_i(\xb_i^{r, t})$ & SG of $f_i(\cdot)$  w.r.t $\xb_{i}$ at iteration $t$, round $r$ \\
		$\eta_i$ & Stepsize of SGD w.r.t $\xb_i$ in round $r$\\
		$\|\cdot\|$ & $l_2$ norm of a vector \\
		$[N]$ &  Set of $\{1, \ldots, N\}$ \\
		\hline\hline
	\end{tabular}
\end{table}

\section{Proposed FedVRA Framework}

In this section,  inspired by the classical distributed ADMM, we propose a unified CVR adaptive FL framework, termed FedVRA, and establish its theoretical property.

\subsection{Algorithmic development}

Let us start from the classical distributed ADMM method by considering the consensus formulation of problem \eqref{eqn: vanilla FL}
\begin{align} \label{eqn: FL prob1}
	\min_{\xb_0, \xb_i,i \in [N]}~ \sum_{i = 1}^{N} \omega_i f_i(\xb_i), ~~{\rm s.t.}&~ \xb_0 = \xb_i,  i \in [N],
\end{align}
where $\xb_i$ is the local model copy owned by client $i$ and $\xb_0$ denotes the global model at the server. Then, we define the corresponding augmented Lagrangian (AL) function
\begin{align} 
	&\Lc(\xb_0, \{\xb_i\}, \{\lambdab_i\}) \triangleq \sum_{i = 1}^{N} \omega_i \Lc_i(\xb_0, \xb_i, \lambdab_i), \label{eqn: AL_function}\\
	&\Lc_i(\xb_0, \xb_i, \lambdab_i)  \triangleq f_i(\xb_i) + \langle \lambdab_i, \xb_0 - \xb_i\rangle + \frac{\gamma_i}{2} \|\xb_0  -\xb_i\|^2,\label{eqn: AL_local_function}
\end{align}where $\lambdab_i$ is the Lagrangian dual variable associated with the equality constraint in  \eqref{eqn: FL prob1}, and $\gamma_i > 0$ is the corresponding penalty parameter. The classical ADMM iteratively optimizes the AL function \eqref{eqn: AL_function} with respect to the variables $\{\xb_i, \lambdab_i, \xb_{0}\}$ in a Gauss-Seidel fashion. This is, for round $r = 0, \ldots$,
\begin{subequations}\label{eqn: update ADMM}
	\begin{align}
		\xb_i^{r+1} =& \arg\min\limits_{\xb_i} \Lc_i(\xb_0^{r}, \xb_i, \lambdab_i^{r}), \forall i \in [N], \label{eqn: xi_ADMM} \\
		\lambdab_i^{r+1} =& \lambdab_i^{r} + 
		\gamma_i(\xb_0^{r} - \xb_i^{r+1}), \forall i \in [N], \label{eqn: dual_ADMM}\\
		\xb_0^{r+1} =& \arg\min\limits_{\xb_0}	\Lc(\xb_{0}, \{\xb_i^{r+1}\},  \{\lambdab_i^{r+1}\}) \notag \\
		=&\beta \sum_{i = 1}^{N} \omega_i  (\gamma_i\xb_i^{r+1} - \lambdab_i^{r+1}) \\
		= & \xb_{0}^r + \beta\sum_{i =1}^{N}\omega_i\gamma_i(\xb_i^{r+1}-\xb_0^r) -\beta \sum_{i = 1}^{N} \omega_i   \lambdab_i^{r+1},\label{eqn: x0_ADMM} 
	\end{align}
\end{subequations}where $\beta \triangleq 1/\sum_{i=1}^{N}\omega_i\gamma_i$.  As seen,  these updates naturally suit for the FL setting as the separable structure of \eqref{eqn: xi_ADMM} and \eqref{eqn: dual_ADMM} over all pairs $\{(\xb_i, \lambdab_i)\}$ enables parallel local update of $(\xb_i, \lambdab_i)$ at the clients while $\xb_0$ is updated by the server.  

\begin{algorithm}[t!] 
	\caption{Proposed FedVRA algorithm} 
\normalsize
	\label{alg: FedVRA}
	\begin{algorithmic}[1]
		\STATE {\bfseries Input:} initial values of $\xb_i^0 = \xb_0$, $\lambdab_i^0 = \lambdab^0=\zerob, \forall i$.
		\FOR{round $r=0$ {\bfseries to} $R - 1$}
		\STATE {\bfseries \underline{Server side:}} sample clients $\Ac^r$ from $[N]$ and broadcast $\xb_0^{r}$.
		\STATE {\bfseries \underline{Client side:}}
		\FOR{client $p = 1$ {\bfseries to} $P$ in parallel}  
		\IF{client $i \notin \Ac^r$} \STATE Set $\xb_i^{r+1} = \xb_0^r, ~\lambdab_i^{r+1} = \lambdab_i^r$.	
		\ELSE 
		\STATE Set $\xb_i^{r, 0} = \xb_0^r$.
		\FOR{$t = 0$ {\bfseries to} $Q_i^r - 1$}
		\STATE $\xb_{i}^{r, t+1} = \xb_i^{r,t} - \eta_i(g_i(\xb_i^{r,t}) - \lambdab_i^{r} + \gamma_i(\xb_i^{r,t} - \xb_0^r))$
		\ENDFOR
		\STATE Set $\xb_i^{r+1} = \xb_i^{r, Q_i^r}$.
		\STATE Compute \colorbox{pink!20!white}{ $\lambdab_i^{r+1} = \lambdab_i^{r} +a_i^r\gamma_i (\xb_0^r - \xb_i^{r+1})$}.
		\STATE Upload $\gamma_i(\xb_i^{r+1} -\xb_0^r)$ and $a_i^r$ to the server.
		\ENDIF
		\ENDFOR
		\STATE {\bfseries \underline{Server side:}} Compute $\lambdab^{r+1}$ and $\xb_0^{r+1}$ by 
		\STATE ~~\colorbox{pink!20!white}{\parbox{21em}{$\lambdab^{r+1} = \lambdab^r + \sum\limits_{i \in \Ac^r} \omega_i a_i^r\gamma_i(\xb_{0}^r - \xb_{i}^{r+1})$ }}
		\STATE ~~\colorbox{pink!21!white}{\parbox{21em}{$\xb_0^{r+1} = \xb_0^r + \beta \sum\limits_{i \in \Ac^r} \omega_i d_{i}^r\gamma_i(\xb_i^{r+1} - \xb_0^r) - \beta \lambdab^{r+1}$.}}
		\ENDFOR
	\end{algorithmic}
\end{algorithm}

To go beyond the ADMM,  like FedAvg,  we incorporate the PCP strategy using client sampling and local SGD for handling subproblem \eqref{eqn: xi_ADMM}.   Moreover,  we allow time-varying HLU where the clients can perform different numbers of SGD updates in each round.
In particular, in round $r$, we let the server sample a small set of $m$ clients $\Ac^r \subset [N]$ with $\Prob(i \in \Ac^r) = p_i^r$ and broadcasts $\xb_0^r$ to all clients.
\begin{itemize}
	\item {\bf Local update}:  each client $i \in \Ac^r$ is asked to take $Q_i^r$ consecutive steps of SGD, i.e. $\xb_i^{r, 0} =\xb_0^r, \xb_i^{r+1} = \xb_i^{r, Q_i^r}$, and $\forall t = 0, \ldots, Q_i^r - 1$, 
	\begin{align}		
		\xb_i^{r, t+1} =& \xb_i^{r,t} - \eta_i(\underbrace{g_i(\xb_i^{r,t}) - \lambdab_i^{r} + \gamma_i(\xb_i^{r,t} - \xb_0^r)}_{\text{SGD of} ~\Lc_i ~\text{w.r.t}~ \xb_i^{r,t}}), \label{eqn: xi_FedVRA}
	\end{align}where $\eta_i > 0$ is the stepsize.  Besides, instead of directly using \eqref{eqn: dual_ADMM}, we introduce an \emph{adaptive dual stepsize} $a_i^r$ for the dual variable $\lambdab_i$, like below
	\begin{align}
		\lambdab_i^{r+1} = \lambdab_i^{r} + 
		a_i^r\gamma_i(\xb_0^{r} - \xb_i^{r+1}), \forall i \in \Ac^r. \label{eqn: dual_FedVRA}
	\end{align}
	The advantage of $a_i^r$ lies in that it enables the fine-grained adaptivity to client heterogeneity as will be discussed in detail in Sec. \ref{sec: ADMM_variance_reduction}. Note that $(\xb_{i}, \lambdab_i)$ are unchanged for non-active clients, i.e., $\xb_{i}^{r+1} = \xb_{0}^r, \lambdab_i^{r+1} = \lambdab_i^r, \forall i \notin \Ac^r$.
	\item {\bf Global aggregation}: after receiving $\xb_i^{r+1} - \xb_{0}^r$ and $ \lambdab_i^{r+1}, i \in \Ac^r$, the server aggregates them to produce the new global model $\xb_0^{r+1}$ via \eqref{eqn: x0_ADMM}.
	Here,  we introduce an \emph{aggregation stepsize} $d_i^r$ in \eqref{eqn: x0_ADMM} as follows
	\begin{align}
		&\xb_{0}^{r+1} 
		=  \xb_{0}^r + \beta\sum_{i=1}^{N}\!\omega_id_i^r\gamma_i(\xb_i^{r+1}\!-\!\xb_0^r)\!-\!\beta \sum_{i = 1}^{N}\! \omega_i   \lambdab_i^{r+1}. \label{eqn: FedVRA_x0_1}
	\end{align}
\end{itemize}
The above steps are summarized in Algorithm \ref{alg: FedVRA}. It is worth noting that, instead of updating $\xb_{0}^{r+1}$ via \eqref{eqn: FedVRA_x0_1},  we split it into two steps in Step 19 and Step 20 of Algorithm \ref{alg: FedVRA}. By the fact that $\lambdab^r = \sum_{i=1}^{N} \omega_i\lambdab_i^{r}, \forall r \geq 0$ and $\xb_{i}^{r+1} = \xb_{0}^r, \forall i \notin \Ac^r$,  one can show that \eqref{eqn: FedVRA_x0_1} and Step 19 and Step 20 are equivalent.  The benefit of doing this splitting is that the client $i$ only needs to upload $a_i^r$ to the server instead of $ \lambdab_i^{r+1}, \forall i \in \Ac^r$; see Step 15 of Algorithm \ref{alg: FedVRA}. Therefore, FedVRA has almost the same communication cost per round as FedAvg and does not double it.  Besides, we allow HLU in Algorithm \ref{alg: FedVRA} by using a time-varying $Q_i^r$ to denote the number of local SGD steps for client $i$ at round $r$. Note that, in practice, the value of $Q_i^r$ depends on the number of local data samples, the mini-batch size and the number of epochs, which are predetermined by the data and computational resources of the client $i$ \cite{FedNova_2020,FedProx_2018}. 

\subsection{ADMM is a client-variance-reduced scheme}
\label{sec: ADMM_variance_reduction}

We remark that the proposed FedVRA algorithm reduces to the federated ADMM  when $d_i^r = a_i^r = 1, \forall i, r$ (see Algorithm \ref{alg: FedADMM}).  As mentioned,  it has been found that the distributed ADMM is inherently robust against to data heterogeneity.  Here, let us show that federated ADMM is in fact a CVR scheme. To the end, we present the following lemma  proved in Appendix \ref{appdix: lem1}.

 \begin{algorithm}[h] 
	\caption{Federated ADMM} 
	\label{alg: FedADMM}{
		\begin{algorithmic}[1]
			\STATE {\bfseries Input:} initial values of $\xb_i^0$, $\xb_0^0$, $\lambdab_i^0, \forall i \in [N]$.
			\FOR{round $r=0$ {\bfseries to} $R - 1$}
			\STATE {\bfseries \underline{Server side:}} sample a subset of clients $\Ac^r \subset [N]$ and broadcast $\xb_0^{r}$ to all clients.
			\STATE {\bfseries \underline{Client side:}}
			\FOR{client $i =1$ {\bfseries to} $N$ in parallel} 
			\IF{$i \notin \Ac^r$}
			\STATE Set $\xb_i^{r+1} = \xb_0^r, \lambdab_i^{r+1} = \lambdab_i^r$.
			\ELSE
			\STATE Set $\xb_i^{r, 0} = \xb_0^r$.
			\FOR{$t = 0$ {\bfseries to} $Q_i^r - 1$}
			\STATE $\xb_{i}^{r, t+1} = \xb_i^{r,t} - \eta_i(g_i(\xb_i^{r,t}) - \lambdab_i^{r} + \gamma_i(\xb_i^{r,t} - \xb_0^r)$.
			\ENDFOR
			\STATE Set $\xb_i^{r+1} = \xb_i^{r, Q_i^r}$.
			\STATE Compute $\lambdab_i^{r+1} = \lambdab_i^{r} +\gamma_i (\xb_0^r - \xb_i^{r+1})$.
			\STATE Upload $\gamma_i\xb_i^{r+1} - \lambdab_i^{r+1}$ to the server.
			\ENDIF
			\ENDFOR
			\STATE {\bfseries \underline{Server side:}} Compute $\lambdab^{r+1}$ and $\xb_0^{r+1}$ by 
			\STATE ~~$\lambdab^{r+1} = \lambdab^r + \sum\limits_{i \in \Ac^r} \omega_i \gamma_i(\xb_{0}^r - \xb_{i}^{r+1})$,
			\STATE ~~$\xb_0^{r+1} = \xb_0^r + \beta \sum\limits_{i \in \Ac^r} \omega_i \gamma_i(\xb_i^{r+1} - \xb_0^r) - \beta \lambdab^{r+1}$.
			\ENDFOR
	\end{algorithmic}}
\end{algorithm}

\begin{Lemma} \label{lem: local_update}
	For any round $r \geq 0$ and client $i \in \Ac^r$, if $\gamma_i\eta_i \leq 1$, it holds that $\forall t = 0, \ldots, Q_i^r - 1$,
	\begin{align}
		\xb_i^{r, t+1}
		\!= &~ (1- \gamma_i\eta_i) (\xb_i^{r, t} - \wt \eta_{i}( g_i(\xb_i^{r,t}) + \gamma_i\beta\lambdab^r - \lambdab_i^{r}))+ \gamma_i\eta_i\bigg(\beta \sum_{j = 1}^{N}\omega_j\gamma_j\xb_j^r\bigg), \label{lem: local_update_xi}\\
		\lambdab_i^{r+1} 
		\!= &\gamma_i\eta_i \wt Q_i^{r} \sum_{t= 0}^{Q_i^r - 1}\frac{b_i^{r, t}}{\|\bb_i^r\|_1} g_i(\xb_i^{r,t}) + (1 - \gamma_i\eta_i \wt Q_i^{r})\lambdab_i^r,\label{lem: local_update_dual}
	\end{align}
	where $\wt \eta_{i} \triangleq \frac{\eta_i}{1-\gamma_i\eta_i}$, $\bb_i^r \triangleq [b_i^{r, 0}, b_i^{r, 1}, \ldots, b_i^{r,Q_i^r-1}]^\top \in \Rbb^{Q_i^r}$, $b_i^{r, t} =(1- \gamma_i\eta_i)^{Q_i^r - 1 - t}$, $\wt Q_i^r = \|\bb_i^r\|_1$, $ \lambdab^r \triangleq \sum_{i=1}^{N}\omega_i\lambdab_i^r$.
\end{Lemma}

Impressively, Lemma \ref{lem: local_update} tells that the federated ADMM is actually a CVR scheme which attempts to reduce the inter-client variance.  
To elaborate this,  firstly by \eqref{lem: local_update_dual},  we notice that $\lambdab_i^{r+1} $ accumulates the historical \emph{normalized averaging SGs} $\sum_{t= 0}^{Q_i^r - 1}\frac{b_i^{r, t}}{\|\bb_i^r\|_1} g_i(\xb_i^{r,t})$. Since $\lambdab^r = \sum_{i=1}^{N} \omega_i\lambdab_i^{r}$,  
$\lambdab^r$ stands for the accumulation of the  \emph{global} normalized averaging SG $\sum_{i=1}^{N} \omega_i \sum_{t= 0}^{Q_i^r - 1}\frac{b_i^{r, t}}{\|\bb_i^r\|_1} g_i(\xb_i^{r,t})$. Thus,  the term $ \gamma_i\beta\lambdab^r - \lambdab_i^{r}$ in \eqref{lem: local_update_xi} is actually a \emph{gradient correction} so that 
$g_i(\xb_i^{r,t}) + \gamma_i\beta\lambdab^r - \lambdab_i^{r}$ approximates the \emph{global SG},  
which shares the same spirit as existing CVR schemes.  Furthermore,  since $\xb_i^{r, t+1}$ in \eqref{lem: local_update_xi} considers the combination with the averaged past model $\beta \sum_{j = 1}^{N}\omega_j\gamma_j\xb_j^r$,  it can further avoid the local model from deviating from the global one.  
Therefore,  we conclude that federated ADMM is in fact a CVR scheme.

When compared with SCAFFOLD,  the proposed FedVRA not only can adopt time-varying HLU (i.e.,  using different $Q_i^r$ for different clients and different rounds) but also is more communication efficient since in SCAFFOLD the client needs to upload two vector variables to the server in contrast to one vector variable and one scalar in FedVRA.

\subsection{Improved adaptability beyond ADMM}
The introduction of the \emph{adaptive dual stepsize} $a_i^r$ and \emph{aggregation stepsize} $d_i^r$ in FedVRA (Step 14 and Step 19-20 in Algorithm \ref{alg: FedVRA}) provide two-fold improvements over the vanilla federated ADMM.  The first is that 
the \emph{adaptive dual stepsize} $a_i^r$ enables the algorithm to have extra flexibility in dealing with client heterogeneity,  which would accelerate the algorithm convergence.  
The second is that the algorithm can flexibly control the biasness of the global model for any client sampling scheme.

To see the impact of $a_i^r$, one can follow the same idea as Lemma \ref{lem: local_update}  to show that  $\xb_i^{r, t+1}$ and $\lambdab_i^{r+1} $ of Algorithm \ref{alg: FedVRA} satisfy
\begin{align}
	&\xb_i^{r, t+1}
	=  (1- \gamma_i\eta_i) (\xb_i^{r, t} - \wt \eta_{i}( g_i(\xb_i^{r,t}) + \gamma_i\beta\lambdab^r - \lambdab_i^{r})) \notag \\
	&~~~~~~~~~~~~~~~+ \gamma_i\eta_i\bigg(\xb_{0}^{r-1} + \beta \sum_{j  \in \Ac^r}\omega_jd_i^r\gamma_j(\xb_j^r - \xb_{0}^{r-1})\bigg), \label{lem: FedVRA_local_update}\\
	&\lambdab_i^{r+1} 
	= a_i^r\gamma_i\eta_i \wt Q_i^{r}G_i^r + (1 - a_i^r\gamma_i\eta_i \wt Q_i^{r})\lambdab_i^r,\label{lem: FedVRA_dual_update}
\end{align}where $G_i^r \triangleq \sum_{t= 0}^{Q_i^r - 1}\frac{b_i^{r, t}}{\|\bb_i^r\|_1} g_i(\xb_i^{r,t})$. 
By comparing \eqref{lem: FedVRA_dual_update} and \eqref{lem: local_update_dual},  one can observe that $a_i^r$ is an independent parameter that controls the weight of current normalized averaging SG $G_i^r$ relative to the historical ones (which are hidden in $\lambdab_i^r$).  The choice of $a_i^r$ is intimately related to the variance-reduction level.  If $a_i^r=0$,  then $\lambdab_i^r=\lambdab^r=0$ and the gradient correction term $\gamma_i\beta\lambdab^r - \lambdab_i^{r}$ vanishes.  Otherwise,  a relative large value of $a_i^r$ is preferred when the data distribution gets more non-i.i.d.  since fresh $G_i^r$ is more effective in reducing the inter-client variance than the old ones.   

The impact of the \emph{aggregation stepsize} $d_i^r$ can be understood as follows.  Suppose that $p_i^r=\frac{m}{N}$ for all $i$ and $r$.  Denote $\wt \xb_i^{r+1}$ as the local model of client $i$ when it is active in round $r$. Then, by Step 20 of Algorithm 1, we have 
\begin{align}
		&\E_{\Ac^r}\bigg[\xb_0^r + \beta \sum_{i \in \Ac^r}\omega_i d_i^r\gamma_i(\xb_{i}^{r+1} - \xb_0^r) -  \beta \lambdab^r\bigg] \notag \\
		= & \xb_0^r + \beta \frac{m}{N} \sum_{i =1}^{N}\omega_i d_i^r\gamma_i(\wt \xb_{i}^{r+1} - \xb_0^r) -  \beta \lambdab^r \label{dir effect0}\\ 
		= & \begin{cases}
			\beta	\sum\limits_{i = 1}^{N} \omega_i\gamma_i \wt \xb_{i}^{r+1}-  \beta \lambdab^r,~{\rm if~}d_i^r=\frac{N}{m}, \forall i, r, \\
			(1 - \frac{m}{N})\xb_0^r +\frac{  m}{N}\bigg(\beta \sum\limits_{i =1}^{N}\omega_i \gamma_i\wt\xb_{i}^{r+1}\bigg) -  \beta \lambdab^r, ~ {\rm if~}d_i^r = 1, \forall i, r,
		\end{cases} \label{dir effect}
\end{align}
\!\!where in \eqref{dir effect}  two values of $d_i^r = 1$ and $d_i^r = \frac{N}{m}$ are considered.
Since $ \lambdab^r$ is an approximation of the global SG,  \eqref{dir effect0} shows that the expected global model is a gradient descent step.
From \eqref{dir effect},  the starting point is the
model average $\sum_{i = 1}^{N} \omega_i\gamma_i \wt \xb_{i}^{r+1}$ when $d_i^r = \frac{N}{m}$,  whereas when, $d_i^r = 1$,  the starting point is a convex combination of the model average and the past global model $\xb_0^r$. Thereby,  choosing a larger value of $d_i^r$ would reduce the bias of the global model and accelerate the algorithm convergence.

\subsection{Convergence analysis}

The following theorem delineates the convergence conditions for the proposed FedVRA algorithm.
The proofs are presented in Appendix \ref{sec: fea_thm1} and \ref{appdix: thm1}.

\begin{Theorem}\label{thm: FedGAPD}
	Let $\Prob(i \in \Ac^r) = p_i^r$ and $0 < \ul p \leq  p_i^r \leq 1, \forall i, r$. Suppose that the parameters $\gamma_i, a_i^r, \eta_i, d_i^r$  $\forall i, r$, satisfy
	\begin{align}
		&\gamma_i \geq \frac{L_i}{2} + \frac{13L_i}{2p_i^r a_i^r\gamma_i\eta_i \wt Q_i^{r}}, \label{thm: condition1} \\
		&\eta_{i} \leq \min\bigg\{\frac{1}{\sqrt{6}\wt Q_i^rL_i}, \frac{1}{\gamma_i}, \frac{1}{(a_i^r+d_i^r)\gamma_i\wt Q_i^r}\bigg\}. \label{thm: condition2} 
	\end{align} Then, under Assumption  \ref{assumption: lower-bounded} and \ref{assumption: SGD_variance}, we have
	\begin{align}
		&\frac{1}{R}\sum_{r = 0}^{R - 1}\E[\|\nabla f(\xb_0^r)\|^2] 
		\leq  	\frac{2D_1(P^0- \ul f)}{\beta R}+\frac{5D_2\sigma^2}{4S}, \label{thm1: FedGAPD_bd}
	\end{align}where $D_1\triangleq \max\limits_{r} \{\sum_{i=1}^{N} \frac{9}{p_i^r(2a_i^r + d_i^r)\gamma_i\eta_i \wt Q_i^{r}}\}$, $D_2 \triangleq \frac{1}{R}\sum_{r = 0}^{R - 1} \sum_{i =1}^{N}\omega_i(D_1(8+ 2p_i^r(2a_i^r + d_i^r)\gamma_i\eta_i \wt Q_i^{r}) + 9p_i^r(a_i^r + d_i^r)\gamma_i\eta_i \wt Q_i^{r})$ and $P^0 \triangleq \E[f(\xb_{0}^0)] + \sum_{i=1}^{N} \omega_i \frac{4\beta \E[\|\nabla f_i(\xb_{0}^0)\|^2]}{p_i^0a_i^0\gamma_i\eta_{i}\wt Q_i^0}$.
\end{Theorem}

Theorem \ref{thm: FedGAPD} shows that, if the mini-batch size $S = \sqrt{R}$, then FedVRA  converges to a stationary solution in the rate $\mathcal{O}(\frac{1}{R}+ \frac{\sigma^2}{\sqrt{R}})$.
Since the analysis does not make any assumption on the data homogeneity and client sampling strategy,  FedVRA is robust to the non-i.i.d.  data distribution and can adapt to arbitrary client sampling schemes.  Besides, since the number of local updates $Q_i^r$ can be different for different clients and communication rounds,  FedVRA is also robust to time-varying {HLU}.  These aspects are novel when compared to the existing FL algorithms (see Table 1).

It is also observed from Theorem 1 that the convergence of FedVRA is influenced by the constants $D_1$ and $D_2$, which are closely related to the stepsizes $a_i^r$ and $d_i^r$. In particular, if $\gamma_i\eta_{i}$ is fixed and $p_i^r\gamma_i\eta_i \wt Q_i^{r} \ll 1$, then increasing $a_i^r$ or $d_i^r$ properly can potentially decrease  both $D_1$ and $D_2$, thereby accelerating the convergence of FedVRA. 
Such property is consistent with our discussions in the previous subsection and will also be verified through numerical experiments; see Fig. \ref{fig: effect_dual_glr} in Sec. \ref{sec: sim1}. 

When FedVRA adopts a uniform client sampling with $p_i^r=\frac{m}{N}$ for all $i$ and $r$,  we have the following corollary.

\begin{Corollary} \label{corly: uniform_conv_rate}
	Given $p_i^r = \frac{m}{N}, Q_i^r = Q \geq 1, \forall i, r$, and $S = \sqrt{R}$, FedVRA has a convergence rate $\mathcal{O}(\frac{N}{mR}+ \frac{N\sigma^2}{m\sqrt{R}})$.
\end{Corollary}

The proof of Corollary \ref{corly: uniform_conv_rate} is presented in Appendix \ref{sec: cor1_proof}.  
As shown in Table \ref{table: com_FL_alg}, SCAFFOLD seems slightly better in terms of convergence rate,  However, FedVRA is more communication-efficient per communication round as discussed in Sec. 2.2.
In addition,  numerical results suggests that FedVRA has a faster convergence behavior than SCAFFOLD; see Fig. 3 of Sec. \ref{sec: sim1}.
\begin{Rmk} {\rm (On the choice of $a_i^r$ and $d_i^r$) Above analysis motivates us to increase the adaptive stepsizes $a_i^r$ and $d_i^r$ to accelerate the algorithm convergence. It would be practically preferred if a particular setup of $(a_i^r, d_i^r)$ is provided. In practice, we suggest choosing these two parameters so that they satisfy $d_i^r = \frac{1}{p_i^r}$ and $(a_i^r + d_i^r)\eta_i^r\gamma_i {\tilde Q_i^r} \leq 1$. As explained below \eqref{dir effect}, the former choice of $d_i^r = \frac{N}{m}$ when $p_i^r=\frac{m}{N}$ reduces the bias of the global model. On the other hand, the latter condition $(a_i^r + d_i^r)\eta_i^r\gamma_i {\tilde Q_i^r} \leq 1$ is inspired by the condition of \eqref{thm: condition2}. Note that the parameters $(a_i^r, d_i^r)$ chosen in Sec. \ref{sec: sim1} satisfy the conditions, and more importantly, such choice yields much faster convergence and better application performance of FedVRA than its counterparts. }
\end{Rmk}

\begin{figure} [t!]
	\centering
	\includegraphics[width=12cm]{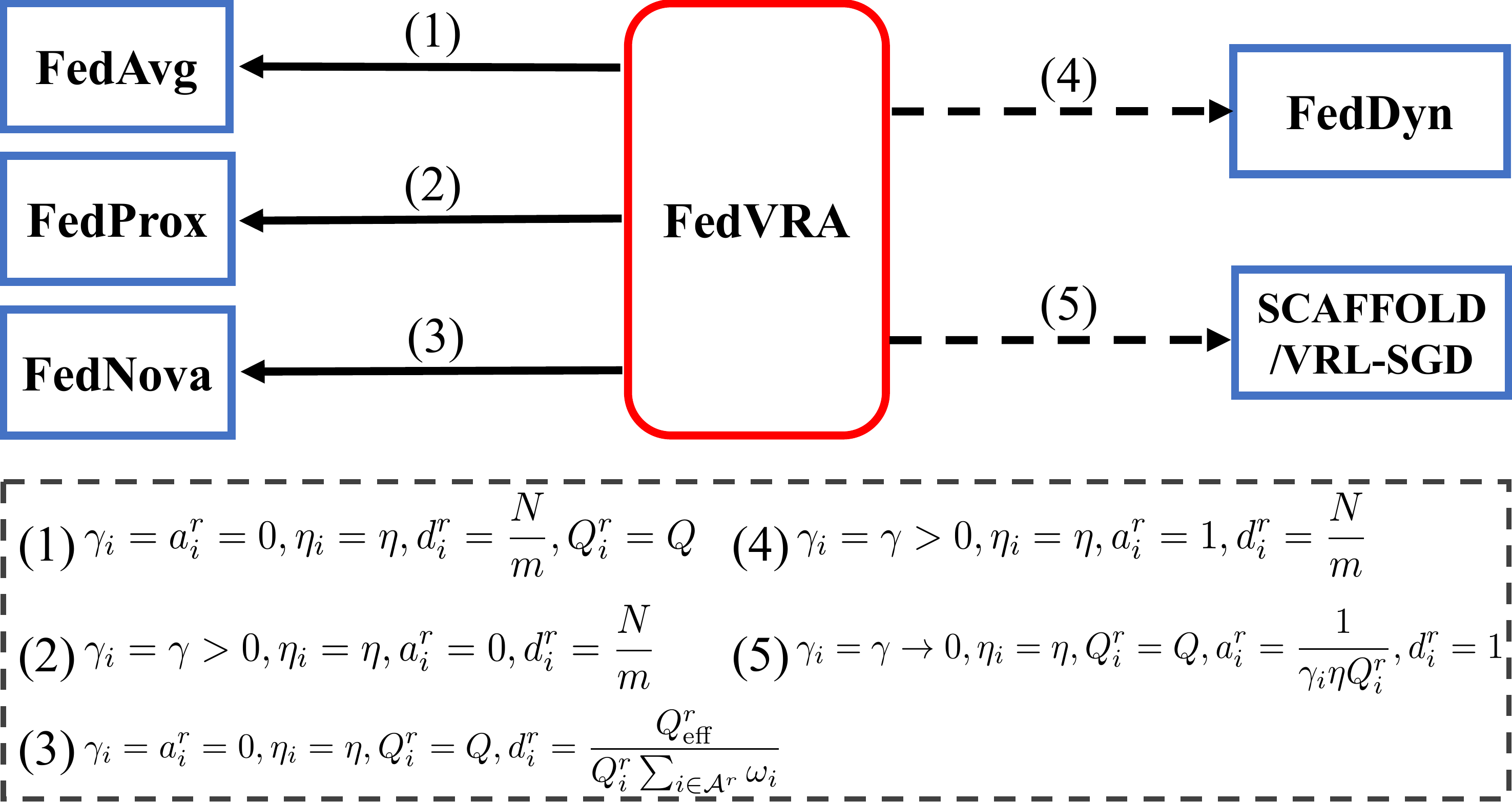} 
	\caption{Connections of FedVRA to existing FL algorithms}
	\label{fig: connection}
\end{figure} 

\section{Connections to Existing FL Algorithms} 

Another advantage of the proposed FedVRA algorithm is that it unifies many FL algorithms, including FedAvg, FedProx, FedNova, SCAFFOLD, VRL-SGD, and FedDyn. These algorithms can be exactly recovered or approximated by FedVRA with specific choices of parameters, as illustrated in Fig. \ref{fig: connection}. This again strengthens our belief of the competitive performance of FedVRA. In particular,
\begin{itemize}
	\item {\bf FedAvg and FedProx:} By simply setting $a_i^r = 0,\eta_{i} = \eta, d_i^r = \frac{N}{m},  Q_i^r = Q$ and $\gamma_i = \gamma = 0, \forall i, r$, then  $\lambdab_i^{r}  = \zerob$ and $b_i^{r,t} =1, \forall r, t$. The update of FedVRA becomes
	\begin{align}
		&\xb_i^{r, t+1} = \xb_i^{r, t} - \eta g_i(\xb_i^{r,t}), \forall i \in \Ac^r,\\
		&\xb_0^{r+1}=\xb_{0}^r +\frac{N}{m}\sum_{i \in \Ac^r} \omega_i(\xb_i^{r+1} - \xb_{0}^r).
	\end{align}which exactly recovers  FedAvg. The former also reduces to FedProx if we keep $\gamma_i = \gamma > 0, i \in [N]$. 
	\item {\bf FedNova:} 
	If we choose $a_i^r = 0, \gamma_i = \gamma = 0, \eta_i = \eta, d_i^r = \frac{Q_{\rm eff}^r}{Q_i^r \sum_{i \in \Ac^r}\omega_i}$, where $Q_{\text{eff}}^r = \frac{\sum_{i \in \Ac^r}\omega_iQ_i^r}{\sum_{i \in \Ac^r}\omega_i}$, the update of $\xb_0$ in FedVRA can be compactly written as
	\begin{align}
		&\xb_0^{r+1} 
		= \xb_0^r -\eta Q_{\text{eff}}^r \sum_{i \in \Ac^r} \frac{\omega_i}{\sum_{i \in \Ac^r} \omega_i}  \sum_{t = 0}^{Q_i^r - 1} \frac{g_i(\xb_i^{r,t})}{Q_i^r}.  \label{eqn: PD_normal}
	\end{align}Therefore, such choice ensures  the equivalence between FedVRA and FedNova \cite{FedNova_2020}. 
	\item {\bf SCAFFOLD/VRL-SGD:} The local update of SCAFFOLD is approximated by choosing $\eta_i = \eta, a_i^r = \frac{1}{\gamma_i\eta Q}, d_i^r = 1, Q_i^r = Q$, and sufficiently small $\gamma_i, \forall i$.  To observe this, let us rewrite \eqref{lem: local_update_xi} and \eqref{lem: local_update_dual} with such choice, which gives that, as $\gamma_i = \gamma \rightarrow 0$, $\forall i \in \Ac^r$,
	\begin{align}
		&\xb_i^{r, t+1} \approx  \xb_i^{r, t} - \eta( g_i(\xb_i^{r,t}) +\lambdab^r - \lambdab_i^{r}), \label{lem: local_update_xi1} \\
		& \lambdab_i^{r+1} \approx \frac{1}{Q}\sum_{t=0}^{Q - 1} g_i(\xb_i^{r,t}).\label{lem: local_update_dual1}
	\end{align}Clearly, the RHS of equations \eqref{lem: local_update_xi1} and \eqref{lem: local_update_dual1} are identical to the local update of SCAFFOLD. When full participation, equation \eqref{lem: local_update_xi1} also reduces to that of VRL-SGD. 
	\item {\bf FedDyn:}  FedDyn is related to FedVRA as the former also optimize the local AL function $\Lc_i$ in \eqref{eqn: AL_function} for the local update. However, FedDyn pursues an exact minimizer to \eqref{eqn: AL_function} for the update of $\xb_{i}^{r+1}$ in place of local SGD. FedVRA approximates FedDyn when we choose $\gamma_i = \gamma > 0, \eta_{i} = \eta, a_i^r = 1, d_i^r = \frac{N}{m}$.
\end{itemize}

\section{Extension to Non-supervised ML problems}

Most FL algorithms are explicitly designed for the smooth problem with one-block of variable, which corresponds to supervised ML tasks. Few  \cite{Shuai_FedMA_2021, Shuai_FedMAJ_2020} aims to solve the following constrained problem with two blocks of variables, which covers many semi-supervised \cite{FedSSL} or unsupervised ML problems \cite{Shuai_SNCPJ_2019}.
\begin{subequations}\label{eqn: Fedprob_twoblock}
	\begin{align}
		\min_{\xb, \{\yb_i\}_i^N}&~ f(\xb, \yb)\triangleq\sum_{i=1}^{N}\omega_i f_i(\xb, \yb_{i})\\
		~~ {\st }&~ \xb \in \Xc, \yb_i \in \Yc_i, i \in [N],
	\end{align}
\end{subequations}where $f_i(\cdot, \cdot)$ is local cost function; $\Xc, \Yc_i$ are closed and convex constraint sets. The variable $\yb_{i}$ corresponds to the pseudo labels of client $i$ in semi-supervised FL \cite{FSSL_Consistency_2021} while it corresponds to the cluster indicators in federated clustering \cite{Shuai_SNCPJ_2019}. It is crucial to handle problem \eqref{eqn: Fedprob_twoblock}  in FL which, however, is much more challenging  to solve than problem \eqref{eqn: vanilla FL} because of the extra block of variable $\yb_i$ and additional constraint sets $\Xc, \Yc_i$. 

We first reformulate  problem  \eqref{eqn: Fedprob_twoblock} to the following consensus form and then apply our previous findings. 
\begin{subequations}\label{eqn: FL_two_consensus}
	\begin{align} 
		\min_{\substack{\xb_i, \xb_0, \yb_i ,\\
				i \in [N]}} ~&\sum_{i = 1}^{N}\omega_i f_i(\xb_i, \yb_i) \\~~{\rm s.t.}~&\xb_{0} \in \Xc, \xb_0 = \xb_i,\yb_i \in \Yc_i , i \in [N]. 
	\end{align}
\end{subequations}Following the same spirit as FedVRA, we present a variant of FedVRA, termed FedVRA-U, to solve problem \eqref{eqn: FL_two_consensus} where the schemes of local SGD and alternating minimization are applied to obtain both the update of $\yb_i$ and $\xb_i$ at each round. The detailed procedure is summarized in Algorithm \ref{alg: FL_PD2}. Note that 
$g_i^x(\xb_i^{r, t})$ denotes the SG of $f_i(\xb_{i}^{r,t}, \yb_{i}^{r+1})$ w.r.t $\xb_{i}$ at iteration $t$ of round $r$ while $g_i^y(\yb_i^{r, t})$ denotes the SG of $f_i(\xb_0^r, \yb_i^{r, t})$  w.r.t. $\yb_{i}$ at iteration $t$ of round $r$. $\Pc_{\Xc}$ and $\Pc_{\Yc_i}$ respectively denote the projection onto the set $\Xc$ and $\Yc_i$. $\eta_i^y$ denotes the stepsize for the update of $\yb_i$.

{\bf Convergence analysis:} The following theorem delineates the convergence result of FedVRA-U. The proof is presented in Appendix \ref{sec: thm2_proof}. Theorem \ref{thm: FedVRA-U} tells that FedVRA-U converges sublinearly to a stationary solution to problem  \eqref{eqn: Fedprob_twoblock}. Similar to FedVRA, it is resilient to heterogeneous clients and arbitrary client sampling schemes, which validates the strong scalability of FedVRA to the challenging FL problem \eqref{eqn: Fedprob_twoblock}. The performance of FedVRA-U will be examined later; See Sec. 5.2 for details.
\begin{Theorem}\label{thm: FedVRA-U}
	Let $\Prob(i \in \Ac^r) = p_i^r$ and $0 < \ul p \leq  p_i^r  \leq \ol p < 1, \forall i, r$. Suppose that the parameters $\gamma_i, \eta_{i}, \eta_i^y, a_i^r, d_i^r$  $\forall i, r$, satisfy
	\begin{align}
		&\gamma_i \geq \frac{L_i}{2} + \frac{9L_i}{p_i^ra_i^r\gamma_i\eta_i \wt Q_i^{r}\sqrt{1-p_i^r}}, \\
		&\eta_{i} \leq \min\bigg\{\frac{1}{\sqrt{6}\wt Q_i^rL_i}, \frac{1}{\gamma_i}, \frac{1}{(a_i^r+d_i^r)\gamma_i\wt Q_i^r}\bigg\},\\
		& \frac{1}{\eta_{i}^{y}} \geq L_i + 4\beta Q_{y_i}^rL_i^2 \bigg(1 + \frac{400}{(a_i^r\gamma_i\eta_i \wt Q_i^{r})^2}\bigg),
	\end{align}where $\wt Q_i^r$ is defined in Theorem \ref{thm: FedGAPD}, and let the mini-batch size $S = \sqrt{R}$. Then, under Assumption  \ref{assumption: lower-bounded} and \ref{assumption: SGD_variance}, FedVRA-U obtains the convergence rate $\mathcal{O}(\frac{1}{R} + \frac{\sigma^2}{\sqrt{R}})$.
\end{Theorem}

\begin{algorithm}[t!] 
	\normalsize
	\caption{Proposed FedVRA-U} 
	\label{alg: FL_PD2} {
		\begin{algorithmic}[1]
			\STATE {\bfseries Input:} initial values of $\xb_0^0 = \xb_1^0 = \ldots, \xb_N^0 $, $\yb_i^0, \lambdab_i^0 = \lambdab= \zerob, \forall i$.
			\FOR{round $r=0$ {\bfseries to} $R - 1$}
			\STATE {\bfseries \underline{Server side:}} sample clients $\Ac^r$ from $[N]$ and broadcast $\xb_0^{r}$.
			\STATE {\bfseries \underline{Client side:}}
			\FOR{client $i=1$ {\bfseries to} $N$ in parallel} 
				\IF{client $i\notin \Ac^r$} \STATE  Set $\xb_i^{r+1} = \xb_0^r, \yb_{i}^{r+1} = \yb_{i}^r, \lambdab_i^{r+1} = \lambdab_i^r$.	
				\ELSE 
			
			\STATE Set $\yb_i^{r, 0} = \yb_i^r$, $\xb_i^{r, 0} = \xb_0^r$.
			\FOR{epoch $t = 0$ {\bfseries to} $Q_{y_i}^r - 1$}
			\STATE \colorbox{pink!20!white}{$\yb_i^{r, t+1} = \Pc_{\Yc_i}(\yb_{i}^{r, t} - \eta_{i}^{y}g_i^y( \yb_i^{r,t}))$}
			\ENDFOR
			\STATE Set $\yb_i^{r+1} = \yb_i^{r, Q_{y_i}^r}$. 
			\FOR{epoch $t = Q_{y_i}^r$ {\bfseries to} $\hat Q_i^r- 1$}
			\STATE \colorbox{pink!20!white}{$\xb_{i}^{r, t+1} = \xb_i^{r,t} - \eta_i(g_i^x(\xb_i^{r,t}) - \lambdab_i^{r} + \gamma_i(\xb_i^{r,t} - \xb_0^r))$}
			\ENDFOR
			\STATE Set $\xb_i^{r+1} = \xb_i^{r, \hat Q_i^r}$.
			\STATE Compute $\lambdab_i^{r+1} = \lambdab_i^{r} + a_i^r \gamma_i(\xb_0^r - \xb_i^{r+1})$.
			\STATE Upload $\gamma_i(\xb_i^{r+1} - \xb_0^r)$ and $a_i^r$ to the server.
				\ENDIF
			\ENDFOR
			\STATE {\bfseries \underline{Server side:}} Compute $\lambdab^{r+1}$ and  $\xb_0^{r+1}$ via 
			\STATE ~~$\lambdab^{r+1} = \lambdab^r + \sum\limits_{i \in \Ac^r} \omega_i a_i^r\gamma_i(\xb_{0}^r - \xb_{i}^{r+1})$
			\STATE ~~\colorbox{pink!20!white}{$\xb_0^{r+1} = \Pc_{\Xc}\big(\xb_0^r + \beta \sum\limits_{i \in \Ac^r} \omega_i d_{i}^r\gamma_i(\xb_i^{r+1} - \xb_0^r) - \beta \lambdab^{r+1}\big)$}.
			\ENDFOR
	\end{algorithmic}}
\end{algorithm}

\section{Experiment Results}
\label{sec: simulation}

In this section, we will examine the performance of the proposed algorithms by comparing them against four baseline FL algorithms, namely, FedAvg \cite{FedAvg_noniid_2019}, FedProx \cite{FedProx_2018}, SCAFFOLD \cite{SCAFFOLD_2020}, and FedDyn \cite{FedDyn_2021}. All presented results are averaged over 5 runs with different and randomly generated initial points.

All experiments were implemented using Pytorch. We reimplemented all baselines by following their respective algorithm descriptions. For methods including FedAvg, FedProx and FedNova, we use an SGD local optimizer with weight decay $10^{-3}$ and no momentum.  For the rest, we implemented the local optimizer based on the SGD optimizer and use the same weight decay. The experiments are performed on a computer equipped with Intel Xeon E5-2680 v4 CPU, 28 GB RAM, 450GB Disk Storage, and NVIDIA RTX A4000 GPU with 16GB memory. Note that the training process mostly runs on the GPU with CUDA version 11.2 and Pytorch version 1.10.

{\bf Datasets and models:} The popular CIFAR-10 \cite{CIFAR10_Krizhevsky09} and MNIST datasets \cite{website_MNIST} are considered for evaluation. These two datasets are widely used in previous FL works and believed to provide convincing experimental results to validate the performance of the proposed FedVRA algorithm. Specifically, the CIFAR-10 dataset contains 50K training  images and 10K test ones while the MNIST dataset has 60K training images of handwritten digits and 10K test ones. 
For the CIFAR-10 dataset, we use data augmentation (random crops, and horizontal flips) and normalize each individual image sample. For the MNIST dataset, we just normalize the image samples. 

We simulate the FL process by distributing the training samples of each dataset to $N = 100$ clients in two ways: {\bf IID} and {\bf Non-IID}. The IID distributed data is generated by randomly assigning training data samples to all clients. To obtain the non-IID distributed data, we follow the heterogeneous data partition method  as in \cite{BN_FL_2019, FedDyn_2021} where each client is allocated data samples of only a few class labels according to the Dirichlet distribution. Note that the Dirichlet parameter is set to $0.2$ by default to ensure that 80\% of each client's local data belong to about 2 classes. Besides, the test set of each dataset is used to evaluate the generalization performance (test accuracy) of the trained model parameter.

We respectively adopt a CNN model for the CIFAR-10 and a fully-connection neural network model for the MNIST image classification task. The former is similar to that in \cite{FedAvg_noniid_2019} which consists of two convolutional layers and two fully connected layers while the later is same to that in \cite{FedDyn_2021}.

{\bf Parameter setting:} We consider the parameter configurations by carefully adjusting them for a fair comparison. In particular, the learning rate $\eta$ is searched from the space $\{0.001, 0.01, 0.1, 1\}$ by applying these choices to FedAvg on the datasets.  Thus, all algorithms take the same learning rate $\eta = 0.01$. The mini-batch size $S$ are respectively set to be $50$ and $32$ for the supervised classification task and the semi-supervised classification tasks. At each communication round, we uniformly sample 10\% of the total clients ($|\Ac^r| = 10$), and choose the number of local epochs at random from $[1, 5]$ if HLU is considered, and otherwise set it to be 2 for each client. Other algorithm specific parameters are tuned individually for each algorithm based on its empirical performance. 
\begin{itemize}
	\item FedProx: we search the regularization parameter $\mu$ from the space $\{0.01, 0.1, 1, 10\}$ and select $\mu = 0.1$ for all cases.
	\item SCAFFOLD: we choose the global learning rate $\eta_g$ as that in \cite{SCAFFOLD_2020}, i.e., $\eta_g = 1$.
	\item FedDyn: we search the penalty parameter $\alpha$ from the space $\{0.01, 0.1, 0.5, 1\}$ and select $\alpha = 0.1$ for all cases.
	\item FedVRA: we decide the choices of $a_i^r$ and $d_i^r$ by searching from the space $\{1, 3, 5, 7, 10, 15, 20\}$. We also choose the parameter $\gamma_i$ from the space $\{0.01, 0.1, 0.5, 1\}$. In particular, for the supervised classification task, we use $\gamma_i = 0.1, \forall i \in [N]$. The parameter $d_i^r$ is set to be $10$ and the parameter $a_i^r$ is set to be $7$ (reps. $10$) for the CIFAR-10 (resp. MNIST) dataset. For the semi-supervised classification task, the parameter $d_i^r = 15$, and the parameters $a_i^r = 7, \gamma_i = 0.1$ (resp. $a_i^r = 5, \gamma_i = 0.5$) on the CIFAR-10 (resp. MNIST) dataset. 
\end{itemize}

\subsection{Evaluation of Algorithm \ref{alg: FedVRA}}
\label{sec: sim1}
\begin{figure} [htb!]
	\centering
	\includegraphics[width=14cm]{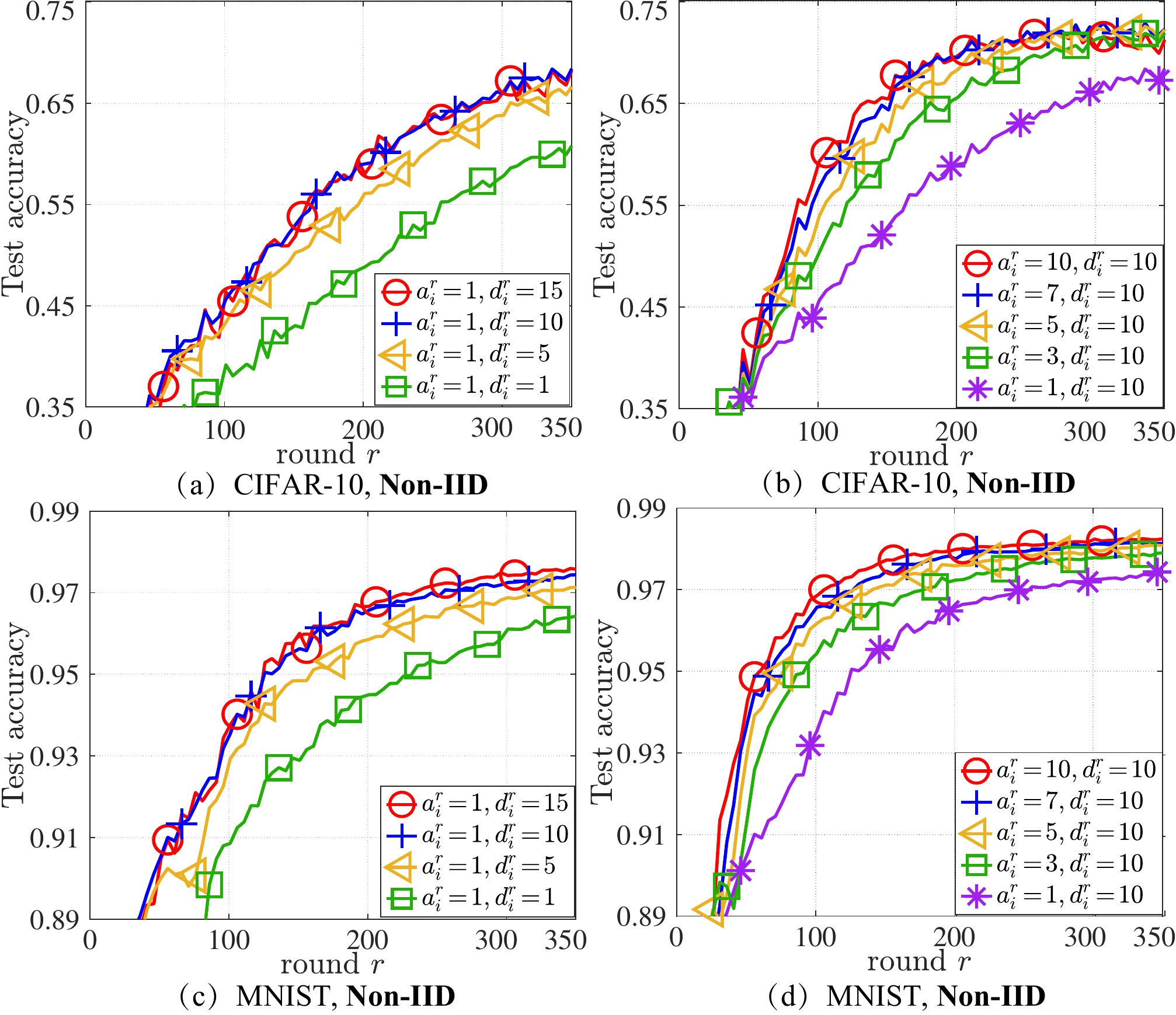}\label{fig: cifar_dual_glr}
	\centering \caption{Performance of FedVRA with different $a_i^r$ and $d_i^r$.}\label{fig: effect_dual_glr}  
\end{figure}

{\bf Effect of $a_i^r$ and $d_i^r$:} In Fig. \ref{fig: effect_dual_glr}, we present the performance of FedVRA with different choices of constant $a_i^r$ and $d_i^r$ on both CIFAR-10 and MNIST datasets.  
One can observe from  Fig. (a) and (c) that for constant $a_i^r$, increasing $d_i^r$ can speed up the convergence but may cause some floors. On the contrary, one can see from Fig. (b) and (d) that under \textbf{Non-IID} data, increasing $a_i^r$ properly can not only speed up the convergence but also achieve a better test performance. The above results corroborate with our analysis that proper values of $a_i^r$ and $d_i^r$ would boost the convergence of FedVRA.  Note that the best choices of $a_i^r$ and $d_i^r$ may depend on the dataset. 

\begin{figure} [htb!]
	\centering
	\includegraphics[width=14cm]{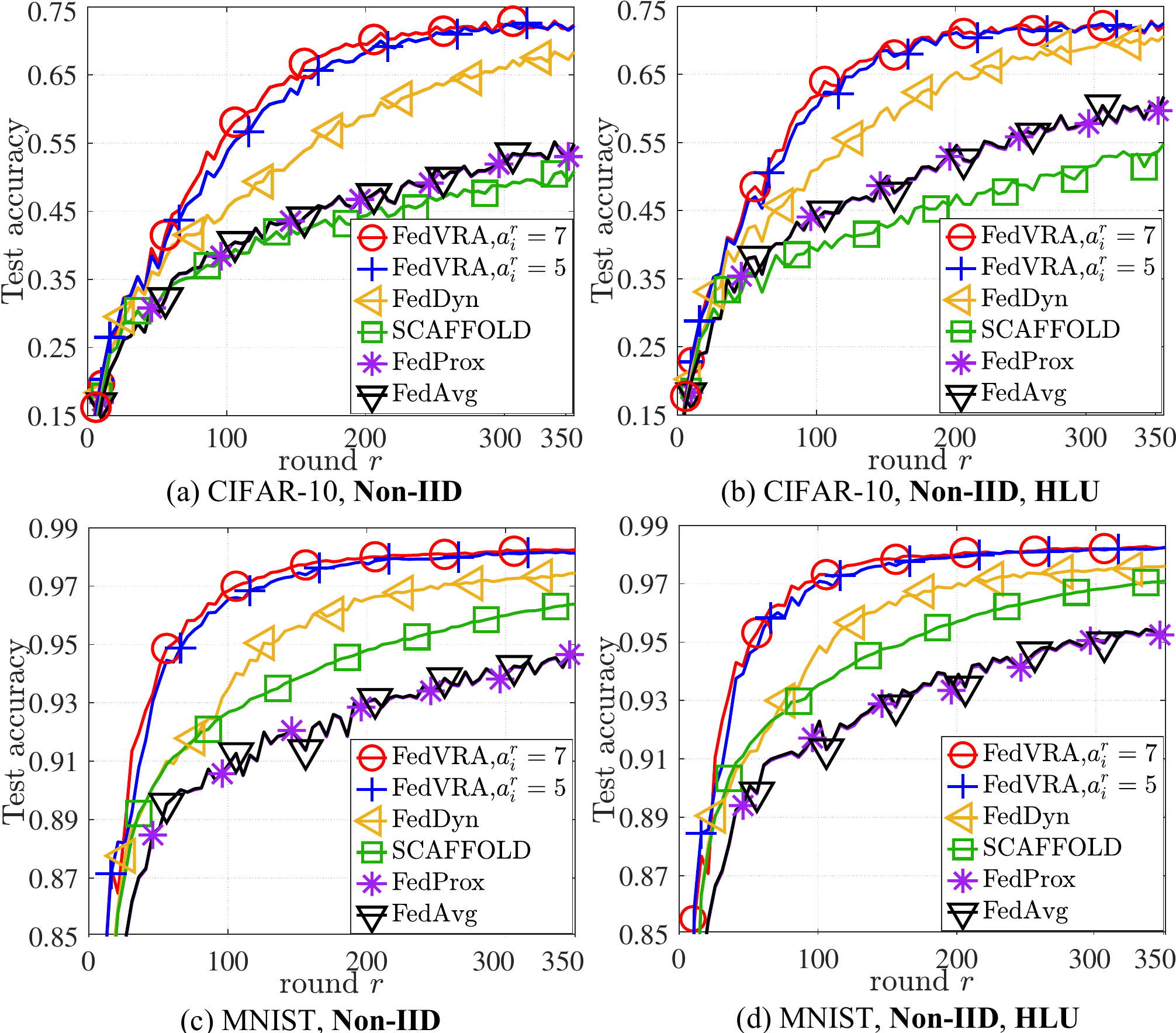}
	\centering \caption{Performance comparison between FedVRA with four FL algorithms. Note that $d_i^r = 10$ for FedVRA.}\label{fig: comparison}  
\end{figure} 

{\bf Performance comparison:} In Fig. \ref{fig: comparison}, we compare FedVRA  with the four FL algorithms on the non-i.i.d. CIFAR-10 and MNIST datasets.  One can see from Fig. 3(b) that on CIFAR-10 dataset, FedVRA significantly outperforms these FL algorithms in terms of both speed and performance.  It is worth noting that SCAFFOLD performs comparably to FedAvg under {\bf Non-IID} and even worse than FedAvg if {\bf HLU} is also applied. But our proposed FedVRA algorithm is more resilient to both non-i.i.d data and HLU, thereby yielding much better performance. We can also observe in Fig. 3(c)(d) the superior performance of FedVRA on the MNIST dataset.

Table \ref{table: comp_cifar} and \ref{table: comp_mnist} respectively summarize the detailed results, including test accuracy achieved and  the number of communication rounds required, on the CIFAR-10 and MNIST datasets. The parameters $d_i^r =10$ for FedVRA and $a_i^r = 7$ (resp. $a_i^r = 10$) for FedVRA on the CIFAR-10 (resp. MNIST) dataset. One can observe that FedVRA performs the best and achieves much higher test accuracy than FedAvg, FedProx and SCAFFOLD.  More importantly, FedVRA is almost 1 time faster than FedDyn and three times faster than the rest. For instance, FedVRA takes $114$ rounds to achieve $\%60$ accuracy on the non-i.i.d CIFAR-10 dataset while FedDyn needs 206 rounds and the rest requires more than $450$ rounds. The same trend is observed in the case of {\bf IID} and {\bf Non-IID + HLU}. While FedVRA seems slightly more sensitive to different initial points than FedDyn as observed from the 3rd row of Table \ref{table: comp_cifar}, its faster convergence and better application performance, especially under Non-IID and HLU cases, demonstrates its superior CVR capability over the others. On the MNIST dataset, it can also be observed from Table \ref{table: comp_mnist} that FedVRA outperforms the rest in terms of convergence speed, test accuracy achieved and even stability to initial points.

\begin{table}[t!]
	\normalsize
	\setlength{\tabcolsep}{6.5mm}
	\caption{Performance comparison between FedVRA and four FL algorithms on the CIFAR-10 dataset. "$\#$R$-$XX" denotes the number of rounds required to reach XX$\%$ accuracy.} 
	\label{table: comp_cifar}
	\begin{tabular}{l|c|c|c|c|c}
		\toprule
		\midrule
		\rowcolor{gray!50} 
		Cases                                                                         & Algorithm & Accuracy & \#R-50 & \#R-60 & \#R-70 \\ \midrule
		& FedAvg    &   67.72  $\pm$ 0.20      &   139     &  282      &   $>$ 500      \\ \cline{2-6} 
		& FedProx   &    67.62 $\pm$ 0.16    &    139    &  284      &    $>$ 500     \\ \cline{2-6} 
		& SCAFFOLD  &   67.32  $\pm$ 0.25     &   143     &    287    &  $>$ 500      \\ \cline{2-6} 
		& FedDyn    &   74.53  $\pm$ 0.14      &    78    &    134    &  250      \\ \cline{2-6} 
		\multirow{-5}{*}{IID}                                                         & FedVRA   &  \textbf{75.47 $\pm$ 0.23 }        &      \textbf{40}  &   \textbf{67}    &  \textbf{121}      \\ \midrule
		& FedAvg    &   61.03   $\pm$ 1.00     &     231   & 477      &   $>$ 500     \\ \cline{2-6} 
		& FedProx   &    61.03   $\pm$ 0.99   &   231     &   459     &  $>$ 500      \\ \cline{2-6} 
		& SCAFFOLD  &    55.49  $\pm$ 0.71    &  318      & $>$ 500       &  $>$ 500      \\ \cline{2-6} 
		& FedDyn    &     71.65  $\pm$ 0.41   &   130     &   206     &  389      \\ \cline{2-6} 
		\multirow{-5}{*}{Non-IID}                                                     & FedVRA   &   \textbf{ 73.70$\pm$ 0.54 }      &   \textbf{ 76}    &     \textbf{114}   &    \textbf{191}    \\ \midrule
		& FedAvg    &  65.20 $\pm$ 1.29       & 150       &    305    &   $>$ 500     \\ \cline{2-6} 
		& FedProx   &   65.08    $\pm$ 1.29   &   157     &  305      &    $>$ 500    \\ \cline{2-6} 
		& SCAFFOLD  &   59.40  $\pm$2.98     &    260    &    481    &  $>$ 500        \\ \cline{2-6} 
		& FedDyn    &  72.19  $\pm$ 0.57      &    95    &  150      &  306      \\ \cline{2-6} 
		\multirow{-5}{*}{\begin{tabular}[c]{@{}c@{}}Non-IID\\ + HLU\end{tabular}} & FedVRA   &   \textbf{73.77$\pm$ 1.02}       &  \textbf{59}      & \textbf{87}       &    \textbf{159}    \\\bottomrule
	\end{tabular} 
\end{table}

\begin{table}[t!]
	\normalsize
	\setlength{\tabcolsep}{6.5mm}
	\caption{Performance comparison between FedVRA and four FL algorithms on the MNIST dataset. "$\#$R$-$XX" denotes the number of rounds required to reach XX$\%$ accuracy.}
	\label{table: comp_mnist}
	\begin{tabular}{l|c|c|c|c|c}
		\toprule
		\midrule
		\rowcolor{gray!50} 
		Cases                                                                         & Algorithm & Accuracy & \#R-90 & \#R-95 &\#R-97  \\ \midrule
		& FedAvg    & 95.64   $\pm$ 0.12 & 63     &   373 & $>$ 500  \\ \cline{2-6} 
		& FedProx   & 95.62   $\pm$ 0.11    &   63    &   393  &  $>$ 500  \\ \cline{2-6} 
		& SCAFFOLD  & 97.04    $\pm$ 0.05      & 44     &     212 &  485    \\ \cline{2-6} 
		& FedDyn    &  97.74    $\pm$ 0.09 &  42   &   125  & 251 \\ \cline{2-6} 
		\multirow{-5}{*}{Non-IID}                                                     & FedVRA   &    \textbf{98.34 $\pm$ 0.06 }      &     \textbf{29}      &    \textbf{60} &   \textbf{107}  \\ \midrule
		& FedAvg    & 96.18 $\pm$ 0.36 & 49  &  273  &   $>$ 500   \\ \cline{2-6} 
		& FedProx   & 96.15    $\pm$ 0.36    &    49    &  280 &  $>$ 500   \\ \cline{2-6} 
		& SCAFFOLD  &  97.59   $\pm$ 0.05    &  33   &  162    &  327 \\ \cline{2-6} 
		& FedDyn    & 97.86  $\pm$ 0.10 & 34 &   106  &  200  \\ \cline{2-6} 
		\multirow{-5}{*}{\begin{tabular}[c]{@{}c@{}}Non-IID\\ + HLU\end{tabular}} & FedVRA   &   \textbf{98.24 $\pm$ 0.04}       &  \textbf{ 24}     &     \textbf{49}  & \textbf{87} \\ \bottomrule
	\end{tabular}
\end{table} 

\begin{figure} [t!]
	\centering
	\includegraphics[width=14cm]{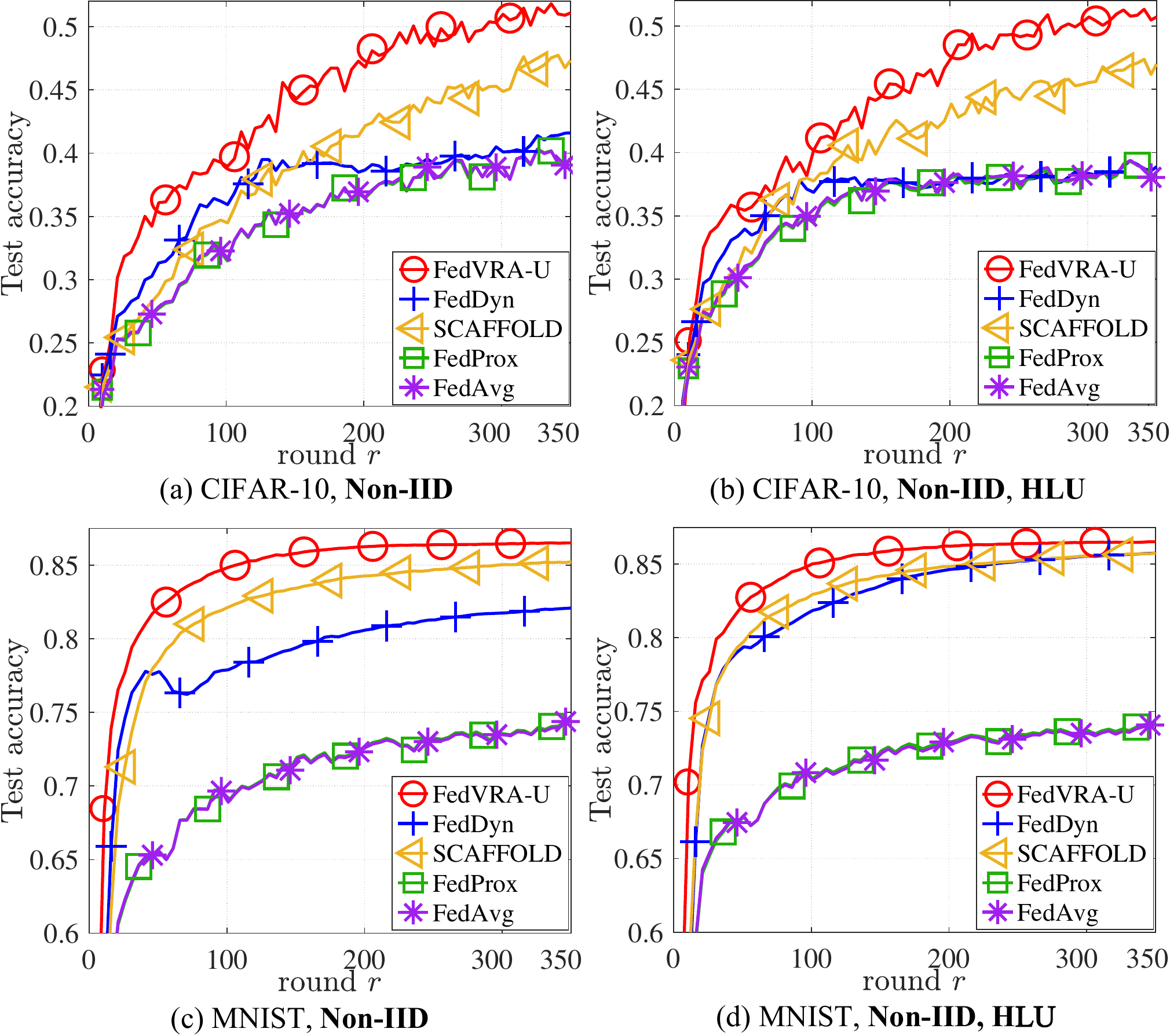}
	\centering \caption{Performance comparison between FedVRA-U with the four FL algorithms.}\label{fig: comparison_ssl} 
\end{figure}  

\subsection{Evaluation of Algorithm \ref{alg: FL_PD2}}
We consider the semi-supervised classification task with the same model as in \cite{FedSSL} on the CIFAR-10 and MNIST datasets. After data distribution, in each client, we randomly select $90\%$ of the local data samples and treat them as the unlabeled data to simulate the semi-supervised FL scenarios where each client only has a few data samples with labels. Then, we examine the performance of FedVRA-U against the four FL algorithm after adapting them to the setting of semi-supervised FL. For FedVRA-U, the parameter $d_i^r = 15$, and the parameters $a_i^r = 7, \gamma_i = 0.1$ (resp. $a_i^r = 5, \gamma_i = 0.5$) on the CIFAR-10 (resp. MNIST) dataset. 

Fig. \ref{fig: comparison_ssl} depicts the performance of FedVRA-U and other FL algorithms under different settings. As seen, the proposed FedVRA-U still performs the best and significantly outperform the rest for all cases. Intriguingly, SCAFFOLD performs better than FedDyn, FedProx and FedAvg, but it performs worse than FedVRA-U. 

\section{Conclusion}\label{sec: conclusion}
In this work, inspired by ADMM, we propose an unified client-variance-reduced adaptive FL framework, FedVRA. As a special instance of FedVRA, federated ADMM is a CVR scheme, which explains its inherent robustness to massive heterogeneous clients. FedVRA adopts two novel adaptive stepsizes, which makes it not only retain the ability of CVR but also enjoy better adaptation to the degree of client heterogeneity and client sampling schemes. In addition, we spotlight that FedVRA unifies many representative FL algorithms. Its superior performance is validated both theoretically and empirically.

\appendices

\section{Proof of Lemma \ref{lem: local_update}}
\label{appdix: lem1}

\noindent{\bf Proof:} According to the update rules of $\xb_i$  and $\xb_0$ in the Algorithm S1, if $i \in \Ac^r$, we have
\begin{align}
	\xb_i^{r, t+1} 
	=& \xb_i^{r, t} - \eta_i( g_i(\xb_i^{r,t}) - \lambdab_i^{r} + \gamma_i(\xb_i^{r,t} -\xb_0^r)) \notag \\
	=&(1- \gamma_i\eta_i) \xb_i^{r, t} - \eta_i( g_i(\xb_i^{r,t}) - \lambdab_i^{r} + \gamma_i(\xb_i^{r,t} -\xb_0^r)) \notag \\
	=&(1- \gamma_i\eta_i) (\xb_i^{r, t} - \wt \eta_{i}( g_i(\xb_i^{r,t}) - \lambdab_i^{r} ))+ \gamma_i\eta_i\xb_0^r \notag \\
	=& (1- \gamma_i\eta_i)(\xb_i^{r, t} - \wt \eta_{i}( g_i(\xb_i^{r,t}) - \lambdab_i^{r})) + \gamma_i\eta_i \beta\sum_{j = 1}^{N}\omega_j(\gamma_j\xb_j^{r} - \lambdab_j^r)  \label{lem: local_update_bd1} \\
	= & (1- \gamma_i\eta_i) (\xb_i^{r, t} - \wt \eta_{i}( g_i(\xb_i^{r,t}) + \gamma_i\beta \lambdab^r - \lambdab_i^{r})) + \gamma_i\eta_i\beta \sum_{j = 1}^{N}\omega_j\gamma_j\xb_j^r, \label{lem: local_update_bd2}
\end{align}where $\wt \eta_{i} \triangleq \frac{\eta_i}{1-\gamma_i\eta_i}, \lambdab^r \triangleq\sum_{i=1}^{N}\omega_i\lambdab_i^r$; \eqref{lem: local_update_bd1} follows because of the update of $\xb_0$ in step 18 of Algorithm S1. Meanwhile, we follow the update rule of $\xb_i$ in step 11 of Algorithm S1 and obtain
\begin{align}
	\xb_i^{r+1} - \xb_0^r 
	=  \xb_i^{r, Q_i^r} - \xb_0^r 	=&(1-\gamma_i\eta_i)(\xb_i^{r, Q_i^r - 1} - \xb_0^r) -  \eta_i( g_i(\xb_i^{r,Q_i^r - 1}) - \lambdab_i^{r}). \label{lem: local_update_bd2_1}
\end{align}
By repeating the above procedure, it holds that 
\begin{align}
	\xb_i^{r+1} - \xb_0^r = - \eta_i\sum_{t = 0}^{Q_i^r - 1} (1-\gamma_i\eta_i^r)^{Q_i^r - 1 - t}(g_i(\xb_i^{r, t})  -\lambdab_i^{r})  =& - \eta_i \wt Q_i^{r} \sum_{t= 0}^{Q_i^r - 1}\frac{b_i^{r, t}}{\|\bb_i^r\|_1} (g_i(\xb_i^{r,t}) - \lambdab_i^r),  \label{lem: local_update_bd3}
\end{align}where 
$\bb_i^r \triangleq [b_i^{r, 0}, b_i^{r, 1}, \ldots, b_i^{r,Q_i^r-1}]^\top \in \Rbb^{Q_i^r}$, $b_i^{r, t} =(1-\gamma_i\eta_i)^{Q_i^r - 1 - t}$, $\wt Q_i^r = \|\bb_i^r\|_1$. 
Then, substituting \eqref{lem: local_update_bd3} into the update rule of $\lambdab_i$ (step 14 of Algorithm S1) gives rise to
\begin{align}
	\lambdab_i^{r+1} 
	= \lambdab_i^r + \gamma_i(\xb_0^r - \xb_i^{r+1}) 
	= & \lambdab_i^r +  \gamma_i\eta_i \wt Q_i^{r} \sum_{t= 0}^{Q_i^r - 1}\frac{b_i^{r, t}}{\|\bb_i^r\|_1} (g_i(\xb_i^{r,t}) - \lambdab_i^r) \notag \\
	= & (1 - \gamma_i\eta_i \wt Q_i^{r})\lambdab_i^r +  \gamma_i\eta_i \wt Q_i^{r} \sum_{t= 0}^{Q_i^r - 1}\frac{b_i^{r, t}}{\|\bb_i^r\|_1} g_i(\xb_i^{r,t}). \label{lem: local_update_bd4}
\end{align} This completes the proof.
\hfill $\blacksquare$

\section{Feasibility of Theorem 1}
\label{sec: fea_thm1}
In order to show the feasibility of Theorem, we need to show that there exists $\gamma_i, a_i^r, \eta_i$, and $d_i^r$ such that both (17) and (18) hold. We first note that
\begin{align}
		\gamma_i &\geq 7 L_i, ~~
		\gamma_i\eta_{i}\wt Q_i^r = 1- (1-\gamma_i\eta_{i})^{Q_i^r} \geq \frac{1}{2}\gamma_i\eta_{i}Q_i^r, \label{eqn: theorem 1 parameter property 2}
\end{align}
according which, we can rewrite the conditions in (17) and (18) as
\begin{align}
		&\left(\gamma_i - \frac{L_i}{2}\right)\gamma_i\eta_{i}  Q_i^r\geq \frac{13L_i}{p_i^r a_i^r}, \label{eqn: theorem 1 condition 1}\\
		&\sqrt{6} \eta_i \tilde Q_i^r L_i \leq 1, \label{eqn: theorem 1 condition 2}\\
		& (1- \gamma_i\eta_{i})^{Q_i^r} \geq 1-\frac{1}{a_i^r + d_i^r}, \label{eqn: theorem 1 condition 3}\\
		& \gamma_i\eta_{i} \leq 1. \label{eqn: theorem 1 condition 4}
\end{align}
As $\gamma_i \geq 7L_i$, it is not difficult to show that \eqref{eqn: theorem 1 condition 2} always hold.
While, for \eqref{eqn: theorem 1 condition 1}, it suffices to show 
\begin{align}
		&\frac{13\gamma_i}{14}\gamma_i\eta_{i}Q_i^r\geq \frac{13L_i}{p_i^r a_i^r}, 
		\iff \gamma_i^2\eta_{i}Q_i^r p_i^ra_i^r\geq 14L_i.	\label{eqn: fea_condition1_1-2}
\end{align}
For \eqref{eqn: theorem 1 condition 2}, by the Tylor expansion, we have $(1 - \gamma_i\eta_{i})^{Q_i^r} \geq 1 - 2\gamma_i\eta_{i}Q_i^r$. Thus,  \eqref{eqn: theorem 1 condition 3} can be equivalently represented by 
\begin{align}
		\gamma_i\eta_{i}Q_i^r \leq \frac{1}{2(a_i^r + d_i^r)}.\label{fea_condition2_1}
\end{align}
Now the feasibility problem of Theorem 1 boils down to finding the feasibility problem of \eqref{eqn: theorem 1 condition 4}, \eqref{eqn: fea_condition1_1-2}, and \eqref{fea_condition2_1}.
The above inequalities are feasible by choosing a sufficiently large $\gamma_i$ and a sufficiently small $\eta_{i}$. For example, based on \eqref{fea_condition2_1}, let $\gamma_i\eta_{i}Q_i^r = \frac{1}{2(a_i^r + d_i^r)}$. Then, we have \eqref{eqn: fea_condition1_1-2} that
\begin{align}
		\gamma_i  \geq \frac{28(a_i^r + d_i^r)L_i}{p_i^ra_i^r}.\label{fea_condition2_2}
\end{align}
Meanwhile, to guarantee \eqref{eqn: theorem 1 condition 4}, one can select proper $a_i^r$ and $d_i^r$ such that $\frac{1}{2(a_i^r + d_i^r)Q_i^r} \leq 1$.
Then, choose $\gamma_i$ satisfying \eqref{fea_condition2_2} and $\eta_{i} = \frac{1}{2(a_i^r + d_i^r)\gamma_i Q_i^r}$,
we can make  conditions (17) and (18) hold and thus Theorem 1 is feasible.
	
\section{Proof of Theorem \ref{thm: FedGAPD}}
\label{appdix: thm1}

In this section, we present the poof of Theorem 1, which states the convergence of FedVRA.

\subsection{Preliminary}
\label{appdix: proof_one_round}
Before delving into the proof, let us introduce some useful terms for ease of presentation. To deal with the randomness incurred by partial client participation, we define the virtual sequence $ \{(\wt \xb_i^{r}, \wt \lambdab_i^{r})\}$ by assuming that all clients are active at round $r$, i.e., $\forall i, 0 \leq t \leq Q_i^r - 1$, $\wt \xb_i^{r, 0} = \xb_0^{r}, \wt \xb_{i}^{r+1}=\wt \xb_i^{r, Q_i^r}$, 
\begin{subequations}
	\begin{align}
		& \wt \xb_i^{r, t+1} =\wt \xb_i^{r, t} - \eta_i( g_i(\wt \xb_i^{r,t}) - \lambdab_i^{r} + \gamma_i(\wt \xb_i^{r,t} -\xb_0^r)) ,\\
		&\wt \lambdab_{i}^{r+1} =  \lambdab_{i}^{r} +  a_i^r\gamma_i(\xb_{0}^r - \wt \xb_i^{r+1}), \label{eqn: def_virtual_dual}
	\end{align}
\end{subequations}
We also define the following additional terms that will be used in our proof.
\begin{align}
	&C_1^{i,r} \triangleq (a_i^r + d_i^r)\gamma_i\eta_i \wt Q_i^{r}, ~C_2^{i,r} \triangleq a_i^r\gamma_i\eta_i \wt Q_i^{r}, \label{def: C2}\\
	&\Xi_i^r \triangleq \E[\| \nabla f_i(\xb_0^r) - \lambdab_i^r\|^2], 
	P^r \triangleq \E[f(\xb_0^r)] + \sum_{i = 1}^{N}\omega_i\frac{4\beta\Xi_i^r}{p_i^rC_2^{i,r}}. \label{def: potential}
\end{align}

\subsection{Overview of the proof}

Theorem 1 is challenging to prove as FedVRA considers both local SGD, HLU and arbitrary client sampling. In order to obtain the convergence property of FedVRA, we follow the a similar analysis framework to \cite{SCAFFOLD_2020} and \cite{NESTT_2016}. In particular, we attempt to build a potential function which descents as $\xb_{i}, \lambdab_i, \xb_{0}$ proceeds in FedVRA  by analyzing the one round progress of the cost function $f$ with respect to $\xb_{0}$. To this end, we develop some new lemmas including Lemma 2, 3 and 4 to to overcome the challenges brought by the novel updates of FedVRA. Then by choosing specific choices of the parameters, the potential function constructed descents properly and we can obtain the desired bounds. Details are presented in the next section.


\subsection{Technical lemmas and their proofs}

In this section, we will present some technical lemmas that will be used in the proof of Theorem 1.

\noindent \fbox{\parbox{1\linewidth}{ 
		\begin{Lemma} \label{lem: effect_dual}
			For any round $r$ and client $i$, if $\eta_i\wt Q_i^rL_i \leq \frac{1}{\sqrt{6}}$, it holds that
			\begin{align}
				&\sum_{t= 0}^{Q_i^r - 1}\frac{b_i^{r,t}}{\|\bb_i^r\|_1}\E[\|\nabla f_i(\xb_0^r)  -g_i(\wt\xb_i^{r,t})\|^2] \leq  \frac{1}{2} \Xi_i^r +  \frac{5\sigma^2}{4S}.
			\end{align}
		\end{Lemma}
}} \newline

\noindent {\bf Proof:} By applying \cite[Lemma 4]{SCAFFOLD_2020},  we have
\begin{small}
	\begin{align}
		\sum_{t= 0}^{Q_i^r - 1}\frac{b_i^{r,t}}{\|\bb_i^r\|_1}\E[\|\nabla f_i(\xb_0^r)  -g_i(\wt\xb_i^{r,t})\|^2] 
		\leq& 	\sum_{t= 0}^{Q_i^r - 1}\frac{b_i^{r,t}}{\|\bb_i^r\|_1}\E[\|\nabla f_i(\xb_0^r)  - \nabla f_i(\wt\xb_i^{r,t})\|^2] + \frac{\sigma^2}{S} \\
		\leq & L_i^2	\sum_{t= 0}^{Q_i^r - 1}\frac{b_i^{r,t}}{\|\bb_i^r\|_1}\E[\|\wt\xb_i^{r,t} - \xb_0^r \|^2] + \frac{\sigma^2}{S}, \label{effect_dual_bd1}
	\end{align}
\end{small}where \eqref{effect_dual_bd1} follows by Assumption 2. Then, we proceed to bound the first term in the RHS of \eqref{effect_dual_bd1}. Specifically, the update of $\xb_i$ in Algorithm 1 gives rise to
\begin{small}
	\begin{align}
		\sum_{t= 0}^{Q_i^r - 1}\frac{b_i^{r,t}}{\|\bb_i^r\|_1}\E[\|\wt\xb_i^{r,t} - \xb_0^r \|^2] 
		\leq & \sum_{t= 0}^{Q_i^r - 1}\frac{b_i^{r,t}}{\|\bb_i^r\|_1} \E\bigg[\bigg\|\eta_i^r \sum_{k= 0}^{t - 1}b_i^{r,k} (g_i(\wt\xb_i^{r,k}) - \lambdab_i^r)\bigg\|^2\bigg] \\
		=& \eta_i^2  \sum_{t= 0}^{Q_i^r - 1}\frac{b_i^{r,t}}{\|\bb_i^r\|_1} \E\bigg[\bigg\|  \sum_{k= 0}^{t - 1}b_i^{r,k} (g_i(\wt\xb_i^{r,k}) - \lambdab_i^r)\bigg\|^2\bigg] \notag \\
		\leq &   \eta_i^2\sum_{t= 0}^{Q_i^r - 1}\frac{b_i^{r,t}}{\|\bb_i^r\|_1}\bigg(\sum_{k= 0}^{t - 1}b_i^{r,k}\bigg)\sum_{k= 0}^{t - 1}b_i^{r,k} \E[\|g_i(\wt\xb_i^{r,k}) - \lambdab_i^r\|^2]\label{effect_dual_bd2},
	\end{align}
\end{small}where \eqref{effect_dual_bd2} follows by the Jensen's Inequality.  Furthermore, note that 
\begin{align}
	\sum_{t= 0}^{Q_i^r - 1}\frac{b_i^{r,t}}{\|\bb_i^r\|_1}\bigg(\sum_{k= 0}^{t - 1}b_i^{r,k}\bigg) 
	\leq & \sum_{t= 0}^{Q_i^r - 1}\frac{b_i^{r,t}}{\|\bb_i^r\|_1}\bigg(\sum_{k= 0}^{Q_i^r - 2}b_i^{r,k}\bigg) 
	\leq  \|\bb_i^r\|_1 - b_i^{r, Q_i^r - 1} 
	\leq   \|\bb_i^r\|_1. \label{effect_dual_bd3}
\end{align}
As a result, we have from \eqref{effect_dual_bd2} that 
\begin{align}
	\sum\limits_{t= 0}^{Q_i^r - 1}\frac{b_i^{r,t}}{\|\bb_i^r\|_1}\E[\|\wt\xb_i^{r,t} - \xb_0^r \|^2]
	\leq  (\eta_i\wt Q_i^r)^2\sum_{t= 0}^{Q_i^r - 1}\frac{b_i^{r,t}}{\|\bb_i^r\|_1}\E[\| g_i(\wt \xb_i^{r,t}) - \lambdab_i^r\|^2]. 
\end{align}
Substituting  it into  \eqref{effect_dual_bd1} yields
\begin{align}
	&\sum_{t= 0}^{Q_i^r - 1}\frac{b_i^{r,t}}{\|\bb_i^r\|_1}\E[\|\nabla f_i(\xb_0^r)  -g_i(\wt\xb_i^{r,t})\|^2] 
	\leq   (\eta_i\wt Q_i^rL_i)^2\Psi_i^r + \frac{\sigma^2}{S}. \label{effect_dual_bd3_1}
\end{align}
On the other hand, we obtain
\begin{align}
	\sum_{t= 0}^{Q_i^r - 1}\frac{b_i^{r,t}}{\|\bb_i^r\|_1}\E[\| g_i(\wt \xb_i^{r,t}) - \lambdab_i^r\|^2]
	\leq & \sum_{t= 0}^{Q_i^r - 1}\frac{b_i^{r,t}}{\|\bb_i^r\|_1}\E[\|\nabla f_i(\wt\xb_i^{r,t}) - \lambdab_i^r\|^2] + \frac{\sigma^2}{S} \label{effect_dual_bd4}\\
	=&\sum_{t= 0}^{Q_i^r - 1}\frac{b_i^{r,t}}{\|\bb_i^r\|_1} \E[\|\nabla f_i(\wt\xb_i^{r,t}) - \nabla f_i(\xb_0^r) + \nabla f_i(\xb_0^r) - \lambdab_i^r\|^2]  + \frac{\sigma^2}{S}\notag \\
	\leq & 2L_i^2\sum_{t= 0}^{Q_i^r - 1}\frac{b_i^{r,t}}{\|\bb_i^r\|_1}\E[\|\wt\xb_i^{r,t} - \xb_0^r\|^2] + 2 \Xi_i^r + \frac{\sigma^2}{S}, \label{effect_dual_bd5} \\
	\leq & 2 (\eta_i\wt Q_i^rL_i)^2\sum_{t= 0}^{Q_i^r - 1}\frac{b_i^{r,t}}{\|\bb_i^r\|_1}\E[\| g_i(\wt \xb_i^{r,t}) - \lambdab_i^r\|^2]+  2 \Xi_i^r + \frac{\sigma^2}{S}, \label{effect_dual_bd6}
\end{align}
where \eqref{effect_dual_bd4} follows by \cite[Lemma 4]{SCAFFOLD_2020} and Assumption 3; \eqref{effect_dual_bd5} holds by the Cauchy–Schwarz Inequality;\eqref{effect_dual_bd6} follows due to \eqref{effect_dual_bd3}.
Rearranging the two sides of \eqref{effect_dual_bd6} gives rise to
\begin{align}
	\Psi_i^r\leq& \frac{1}{1- 2 (\eta_i\wt Q_i^rL_i)^2}\bigg(2 \Xi_i^r +   \frac{\sigma^2}{S}\bigg) 
	\leq  \frac{1}{4(\eta_i\wt Q_i^rL_i)^2}\bigg(2 \Xi_i^r +   \frac{\sigma^2}{S}\bigg), \label{effect_dual_bd7}
\end{align}
where \eqref{effect_dual_bd7} follows because $1 - 2(\eta_i\wt Q_i^rL_i )^2 \geq 4 (\eta_i\wt Q_i^rL_i )^2$. Substituting \eqref{effect_dual_bd7} into  \eqref{effect_dual_bd3_1} yields
\begin{align}
	&\sum_{t= 0}^{Q_i^r - 1}\frac{b_i^{r,t}}{\|\bb_i^r\|_1}\E[\|\nabla f_i(\xb_0^r)  -g_i(\wt\xb_i^{r,t})\|^2] \leq  \frac{1}{2} \Xi_i^r +  \frac{5\sigma^2}{4S}.
\end{align}
This completes the proof. 
\hfill $\blacksquare$
\newline

\noindent \fbox{\parbox{1\linewidth}{ 
		\begin{Lemma} \label{lem: opt_progress}
			For any round $r$ and client $i$, it holds that 
			\begin{small}
				\begin{align}
					\Xi_i^{r+1} - \Xi_i^r  
					\leq &\bigg(1 + \frac{4}{p_i^rC_2^{i,r}}\bigg)L_i^2 \E[\|\xb_0^{r+1}-\xb_0^r\|^2]- \frac{p_i^rC_2^{i,r}(p_i^rC_2^{i,r} + 2)}{8}\Xi_i^r+\frac{5\sigma^2p_i^{r}C_2^{i,r}}{4S}\bigg(1 +  \frac{p_i^rC_2^{i,r}}{4}\bigg). \label{lem: opt_progress_bd}
				\end{align} 
			\end{small}
\end{Lemma}}}
\newline

\noindent {\bf Proof:} 	By the definition of $\Xi_i^r$, we have
\begin{align}
	\Xi_i^{r+1}
	=& \E[\|\nabla f_i(\xb_0^{r+1})-\nabla f_i(\xb_0^r) + \nabla f_i(\xb_0^r) - \lambdab_i^{r+1}\|^2]	\notag \\
	\leq &(1 + 1/\epsilon_i) \E[\|\nabla f_i(\xb_0^{r+1})-\nabla f_i(\xb_0^r) \|^2] +(1 + \epsilon_i)\E[\| \nabla f_i(\xb_0^r) - \lambdab_i^{r+1}\|^2] \label{lem: opt_bd0}\\
	= &(1 + 1/\epsilon_i)L_i^2 \E[\|\xb_0^{r+1}-\xb_0^r\|^2]+(1 + \epsilon_i)(1-p_i^r) \Xi_i^r +(1 + \epsilon_i)p_i^r \E[\| \nabla f_i(\xb_0^r) - \wt \lambdab_i^{r+1}\|^2], \label{lem: opt_bd1}
\end{align}where $\epsilon_i > 0$ is a constant, and \eqref{lem: opt_bd0} follows by the fact that $(z_1 + z_2)^2 \leq (1 + \frac{1}{c}) z_1^2 + (1 + c)z_2^2, \forall c > 0$.  Then, by using \eqref{eqn: def_virtual_dual}, we can bound $\E[\| \nabla f_i(\xb_0^r) - \wt \lambdab_i^{r+1}\|_2^2] $ by
\begin{small}
	\begin{align}
		\E[\| \nabla f_i(\xb_0^r) - \wt \lambdab_i^{r+1}\|^2]
		=&\E\bigg[\bigg\| \nabla f_i(\xb_0^r) - (1-C_2^{i,r}) \lambdab_i^{r}-C_2^{i,r}\sum_{t= 0}^{Q_i^r - 1}\frac{b_i^{r,t}}{\|\bb_i^r\|_1} g_i(\wt \xb_i^{r,t})\bigg\|^2\bigg]\notag \\
		= & \E\bigg[\bigg\|(1-C_2^{i,r})( \nabla f_i(\xb_0^r) - \lambdab_i^{r})+C_2^{i,r}\sum_{t= 0}^{Q_i^r - 1}\frac{b_i^{r,t}}{\|\bb_i^r\|_1}( \nabla f_i(\xb_0^r)  -g_i(\wt \xb_i^{r,t}))\bigg\|^2\bigg]\notag \\
		\leq & (1-C_2^{i,r})\Xi_i^r+C_2^{i,r}\sum_{t= 0}^{Q_i^r - 1}\frac{b_i^{r,t}}{\|\bb_i^r\|_1}\E[\|\nabla f_i(\xb_0^r)  -g_i(\wt \xb_i^{r,t})\|^2] \label{lem: opt_bd2}\\
		\leq &\bigg(1-\frac{1}{2}C_2^{i,r}\bigg)\Xi_i^r + \frac{5\sigma^2C_2^{i,r}}{4S}, \label{lem: opt_bd3}
	\end{align}
\end{small}where  \eqref{lem: opt_bd2} follows by the convexity of $\|\cdot\|^2$ and \eqref{lem: opt_bd3} holds thanks to Lemma \ref{lem: effect_dual}. Substituting \eqref{lem: opt_bd3} into \eqref{lem: opt_bd1} yields
\begin{align}
	&\Xi_i^{r+1} - \Xi_i^r  \notag \\
	\leq &(1 + 1/\epsilon_i)L_i^2 \E[\|\xb_0^{r+1}-\xb_0^r\|_2^2] +\frac{5\sigma^2(1 + \epsilon_i)p_i^{r}C_2^{i,r}}{4S}+((1 + \epsilon_i) (1-p_i^rC_2^{i,r}/2)-1)\Xi_i^r. \label{lem: opt_bd4}
\end{align}Let us pick $\epsilon_i = \frac{p_i^rC_2^{i,r}}{4}$, then $(1 + \epsilon_i)(1 - p_i^rC_2^{i,r}/2) -1 = - \frac{p_i^rC_2^{i,r}(p_i^rC_2^{i,r} + 2)}{8}$, and thus \eqref{lem: opt_bd4} can be simplified to	
\begin{align}
	&	\Xi_i^{r+1} - \Xi_i^r  \notag \\
	\leq &\bigg(1 + \frac{4}{p_i^rC_2^{i,r}}\bigg)L_i^2 \E[\|\xb_0^{r+1}-\xb_0^r\|^2]- \frac{p_i^rC_2^{i,r}(p_i^rC_2^{i,r} + 2)}{8}\Xi_i^r+\frac{5\sigma^2p_i^{r}C_2^{i,r}}{4S}\bigg(1 +  \frac{p_i^rC_2^{i,r}}{4}\bigg). \label{lem: opt_bd5}
\end{align} 
This completes the proof.
\hfill $\blacksquare$
\newline

\noindent \fbox{\parbox{1\linewidth}{ 
		\begin{Lemma} \label{lem: one_round_obj}
			For any round $r$ and client $i$, it holds that
			\begin{align}
				\E[f(\xb_0^{r+1})] - \E[f(\xb_0^r)] 
				\leq &  \beta \sum_{i =1}^{N}\omega_i \bigg(1-\frac{1}{2}p_i^rC_1^{i,r}\bigg)\Xi_i^r  +\frac{5\beta \sigma^2}{4S} \sum_{i =1}^{N}\omega_i p_i^rC_1^{i,r}\notag \\
				&-\bigg(\frac{3}{4\beta} -\sum_{i = 1}^{N} \omega_i  \frac{L_i}{2} \bigg)\E[\|\xb_0^{r+1} - \xb_0^r\|^2].\label{lem: one_round_progress_bd}
			\end{align}
\end{Lemma}}}
\newline

\noindent {\bf Proof:} By the definition of $f(\xb)$, we have
\begin{align}
	&\E[f(\xb_0^{r+1}) - f(\xb_0^r)] \notag \\
	= &\sum_{i = 1}^{N} \omega_i \E[f_i(\xb_{0}^{r+1}) - f_i(\xb_{0}^{r})]\notag \\
	\leq & \sum_{i = 1}^{N} \omega_i\E[ \langle \nabla f_i(\xb_0^r), \xb_0^{r+1} - \xb_0^r\rangle] +\frac{1}{2}\sum_{i=1}^{N}\omega_i L_i \E[\|\xb_0^{r+1} - \xb_0^r\|^2] \label{lem: one_round_obj_bd1}\\
	= & \E[\langle\nabla f(\xb_0^r),\xb_0^{r+1} -\xb_0^r\rangle]+\frac{1}{2}\sum_{i=1}^{N}\omega_i L_i \E[\|\xb_0^{r+1} - \xb_0^r\|^2]\label{lem: one_round_obj_bd2} \\
	= & \E[\langle\nabla f(\xb_0^r) + \frac{1}{\beta}(\xb_0^{r+1} - \xb_0^r) , \xb_0^{r+1} - \xb_0^r \rangle] -\sum_{i = 1}^{N} \omega_i \bigg(\frac{1}{\beta} - \frac{L_i}{2} \bigg)\E[\|\xb_0^{r+1} - \xb_0^r\|^2] \\
	\leq &  \beta \E\bigg[\bigg\|\nabla f(\xb_0^r) + \frac{1}{\beta}(\xb_0^{r+1} - \xb_0^r)\bigg\|^2\bigg] -\bigg(\frac{3}{4\beta} -\sum_{i = 1}^{N} \omega_i  \frac{L_i}{2} \bigg)\E[\|\xb_0^{r+1} - \xb_0^r\|^2], \label{lem1: bd1}
\end{align}where \eqref{lem: one_round_obj_bd1} follows by Assumption 2 and \eqref{lem1: bd1} follows because $\langle \zb_1, \zb_2\rangle  \leq \beta \|\zb_1\|^2 + \frac{1}{4\beta}\|\zb_2\|^2, \forall \zb_1, \zb_2$. Then, we proceed to bound the right hand side (RHS) of \eqref{lem: one_round_obj_bd1}. In particular, owing to the update rule of $\xb_{0}$, we get
\begin{align}
	&\E\bigg[\bigg\|\nabla f(\xb_0^r) + \frac{1}{\beta}(\xb_0^{r+1} - \xb_0^r)\bigg\|^2\bigg] \notag \\
	= &	\E\bigg[\bigg\|\sum_{i = 1}^{N}\omega_i\nabla f_i(\xb_0^r) + \sum_{i = 1}^{N}\omega_id_i^r\gamma_i(\xb_i^{r+1} - \xb_0^r) - \sum_{i = 1}^{N}\omega_i\lambdab_i^{r+1}\bigg\|^2\bigg] \notag \\
	\leq & \sum_{i =1}^{N}\omega_i \E[\|\nabla f_i(\xb_0^r)+ d_i^r\gamma_i(\xb_i^{r+1} - \xb_0^r)-\lambdab_i^{r+1}\|^2]  \label{lem1: bd1_1}\\
	= & \sum_{i =1}^{N}\omega_i( p_i^r\E[\|\nabla f_i(\xb_0^r)+d_i^r\gamma_i(\wt\xb_i^{r+1} - \xb_0^r)-\wt\lambdab_i^{r+1}\|^2]+(1-p_i^r)\Xi_i^r), \label{lem1: bd2}
\end{align}where  \eqref{lem1: bd1_1} holds by the convexity of $\|\|^2$ and  \eqref{lem1: bd2} follows because $\Prob(i \in \Ac^r) = p_i^r$. Furthermore, note that
\begin{align}
	&\E[\|\nabla f_i(\xb_0^r)+d_i^r\gamma_i(\wt\xb_i^{r+1} - \xb_0^r)-\wt\lambdab_i^{r+1}\|^2]\notag \\
	= & \E[\|\nabla f_i(\xb_0^r)-\lambdab_i^{r}-C_1^{i,r} \sum_{t= 0}^{Q_i^r - 1}\frac{b_i^{r,t}}{\|\bb_i^r\|_1} (g_i(\wt \xb_i^{r,t}) - \lambdab_i^r)\|^2] \\
	= & \E[\|(1- C_1^{i,r} )(\nabla f_i(\xb_0^r)-\lambdab_i^{r})-C_1^{i,r} \sum_{t= 0}^{Q_i^r - 1}\frac{b_i^{r,t}}{\|\bb_i^r\|_1} (\nabla f_i(\xb_0^r) - g_i(\wt\xb_i^{r,t}))\|^2] \\
	\leq & (1- C_1^{i,r} ) \Xi_i^r+ C_1^{i,r}\sum_{t= 0}^{Q_i^r - 1}\frac{b_i^{r,t}}{\|\bb_i^r\|_1}\E[\|\nabla f_i(\xb_0^r) - g_i(\wt\xb_i^{r,t})\|^2] \label{lem1: bd3} \\
	\leq & \bigg(1- \frac{1}{2}C_1^{i,r} \bigg)\Xi_i^r + \frac{5\sigma^2 C_1^{i,r}}{4S}, \label{lem1: bd4} 
\end{align}where \eqref{lem1: bd3} follows by the convexity of $\|\|^2$ and the fact that $C_1^{i,r} \leq 1$; \eqref{lem1: bd4} holds thanks to Lemma \ref{lem: effect_dual}. As a result, we have from \eqref{lem1: bd2} that 
\begin{align}
	\E\bigg[\bigg\|\nabla f(\xb_0^r) + \frac{1}{\beta}(\xb_0^{r+1} - \xb_0^r)\bigg\|^2\bigg] 
	\leq &\sum_{i =1}^{N}\omega_i\bigg( \bigg(1-\frac{1}{2}p_i^rC_1^{i,r}\bigg)\Xi_i^r +\frac{5\sigma^2p_i^rC_1^{i,r}}{4S}\bigg).\label{lem1: bd5} 
\end{align}
Lastly, substituting \eqref{lem1: bd5}  into \eqref{lem: one_round_obj_bd2} yields
\begin{align}
	&\E[f(\xb_0^{r+1}) - f(\xb_0^r)] \notag \\
	\leq &  \beta \sum_{i =1}^{N}\omega_i \bigg(1-\frac{1}{2}p_i^rC_1^{i,r}\bigg)\Xi_i^r  +\frac{5\beta \sigma^2}{4S} \sum_{i =1}^{N}\omega_i p_i^rC_1^{i,r}-\bigg(\frac{3}{4\beta} -\sum_{i = 1}^{N} \omega_i  \frac{L_i}{2} \bigg)\E[\|\xb_0^{r+1} - \xb_0^r\|^2].
\end{align}
This completes the proof.
\hfill $\blacksquare$

\subsection{Main proof of Theorem 1}

\noindent {\bf Proof:}  we start by combining Lemma \ref{lem: one_round_obj} and Lemma \ref{lem: opt_progress}. In particular, by multiplying the two sides of \eqref{lem: opt_progress_bd} by $\frac{4\beta}{p_i^rC_2^{i,r}}$, taking average w.r.t all clients, and then adding it to \eqref{lem: one_round_progress_bd}, we have
\begin{align}
	P^{r+1} - P^r 
	\leq &  -\frac{\beta}{2}\sum_{i=1}^{N}\omega_ip_i^r(C_1^{i,r} + C_2^{i,r})\Xi_i^r -\sum_{i=1}^{N}\omega_iD_0^{i,r}\E[\|\xb_0^{r+1} - \xb_0^r\|^2] \notag \\
	&+\frac{5\beta \sigma^2}{S} \sum_{i =1}^{N}\omega_i\bigg(1+ \frac{p_i^r(C_1^{i,r} + C_2^{i,r})}{4}\bigg),	\label{thm: descent1}
\end{align}where $D_0^{i,r} \triangleq \frac{3}{4\beta} - \frac{4\beta }{p_i^rC_2^{i,r}}\bigg(1 + \frac{4}{p_i^rC_2^{i,r}}\bigg)L_i^2-\frac{L_i}{2}$.
Next, we claim that $D_0^{i,r}\geq \frac{1}{4\beta}$. To prove it, it suffices to show that $j = \arg\min_i \gamma_i$, $1-  \frac{L_j}{\gamma_j} - \frac{8L_j^2}{\gamma_j^2p_j^rC_2^{j,r}}\bigg(1 + \frac{4}{p_j^rC_2^{j,r}}\bigg)   \geq  0$, which holds if 
\begin{align}
	1-  \frac{L_j}{\gamma_j} - \frac{40L_j^2}{\gamma_j^2(p_j^rC_2^{j,r})^2}  \geq  0,  \forall j, r, \label{thm: condition_D}
\end{align} because $p_j^rC_2^{j,r} \leq 1$.
By finding the root of the quadratic inequality \eqref{thm: condition_D}, we need $\gamma_j \geq \frac{L_j}{2} + \frac{L_j}{2}\sqrt{1+ \frac{160}{(p_j^rC_2^{j,r})^2}}$, which certainly holds owing to the fact that $\gamma_i \geq \frac{L_i}{2} + \frac{13L_i}{2p_i^rC_2^{i,r}}, \forall i, r$. Then, rearranging the two sides of \eqref{thm: descent1} gives rise to
\begin{align}
	&	\sum_{i=1}^{N}\omega_ip_i^r(C_1^{i,r} + C_2^{i,r})\Xi_i^r +\frac{1}{2\beta^2 }\E[\|\xb_0^{r+1} - \xb_0^r\|^2] \notag \\
	\leq & 	\frac{2(P^r - P^{r+1} )}{\beta }+\frac{10 \sigma^2}{S} \sum_{i =1}^{N}\omega_i\bigg(1+ \frac{p_i^r(C_1^{i,r} + C_2^{i,r})}{4}\bigg).	\label{thm: descent2}
\end{align}
Meanwhile, the optimality gap $\|\nabla f(\xb_{0}^r)\|^2$ can be bounded by
\begin{align}
	&\E[\|\nabla f(\xb_{0}^r)\|^2] \notag \\
	= &	\frac{1}{\beta^2}\E[\|\xb_0^r - \xb_0^{r+1} + \xb_0^{r+1} - \xb_0^r + \beta \nabla f(\xb_{0}^r)\|^2] \notag \\
	\leq &  \frac{1 + c}{\beta^2}\E[\|\xb_0^r - \xb_0^{r+1} \|^2] + \frac{1 + 1/ c}{\beta^2}\E[\|\xb_0^{r+1} - \xb_0^r + \beta \nabla f(\xb_{0}^r)\|^2] \label{thm: descent2_1}\\
	\leq &  \frac{1 + c}{\beta^2}\E[\|\xb_0^r - \xb_0^{r+1} \|^2] + \frac{1 + 1/ c}{2}\sum_{i =1}^{N}\omega_i\bigg((2-p_i^rC_1^{i,r})\Xi_i^r +\frac{5\sigma^2p_i^rC_1^{i,r}}{2S}\bigg) \label{thm: descent2_2}\\
	= &  \frac{1 + c}{\beta^2}\E[\|\xb_0^r - \xb_0^{r+1} \|^2]+ \frac{1 + 1/ c}{2}\sum_{i =1}^{N}\omega_i(2-p_i^rC_1^{i,r})\Xi_i^r + \frac{5(1 + 1/c) \sigma^2}{4S} \sum_{i =1}^{N}\omega_ip_i^rC_1^{i,r},	\label{thm: descent3}
\end{align}where \eqref{thm: descent2_1} follows by the fact that $(z_1 + z_2)^2 \leq (1 + \frac{1}{c}) z_1^2 + (1 + c)z_2^2, \forall c > 0$; \eqref{thm: descent2_2} holds by \eqref{lem1: bd5}. Combing \eqref{thm: descent2} and \eqref{thm: descent3} yields
\begin{align}
	&\E[\|\nabla f(\xb_{0}^r)\|^2] \notag \\
	\leq & 	\frac{2D_0^{r}(P^r - P^{r+1} )}{\beta }+\frac{10D_0^{r} \sigma^2}{S} \sum_{i =1}^{N}\omega_i\bigg(1+ \frac{p_i^r(C_1^{i,r} + C_2^{i,r})}{4}\bigg) + \frac{5 (1+1/c)\sigma^2}{4S} \sum_{i =1}^{N}\omega_ip_i^rC_1^{i,r},\label{thm: descent4}
\end{align}
where $D_0^r \triangleq 2(1 + c) + \frac{1 + 1/c}{2}\sum_{i=1}^{N} \frac{2-p_i^rC_1^r}{p_i^r(C_1^{i,r} + C_2^{i,r})} = \sum_{i=1}^{N} \frac{1 + 1/c}{p_i^r(C_1^{i,r} + C_2^{i,r})} + 2(1+c) - \frac{1 + 1/c}{2}\sum_{i=1}^{N} \frac{C_1^r}{C_1^{i,r} + C_2^{i,r}}$.
Let $c = \frac{1}{8}$, then we have
\begin{align}
	2(1+c) - \frac{1 + 1/c}{2}\sum_{i=1}^{N} \frac{C_1^r}{C_1^{i,r} + C_2^{i,r}} = \sum_{i = 1}^{N} \bigg(\frac{9}{4N} - \frac{9C_1^{i,r}}{2(C_1^{i,r} + C_2^{i,r})}\bigg) 
	\leq \frac{9}{4}\sum_{i=1}^{N} \bigg(1 - \frac{2C_1^{i,r}}{C_1^{i,r} + C_2^{i,r}}\bigg) 
	\leq  0,\notag 
\end{align}
which implies that $D_0^r \leq \sum_{i=1}^{N} \frac{9}{p_i^r(C_1^{i,r} + C_2^{i,r})}$. Substituting the result into the RHS of \eqref{thm: descent4} yields
\begin{align}
	&\E[\|\nabla f(\xb_{0}^r)\|^2] \notag\\
	\leq & 	\frac{2D_1^{r}(P^r - P^{r+1} )}{\beta }+\frac{10D_1^{r} \sigma^2}{S} \sum_{i =1}^{N}\omega_i\bigg(1+ \frac{p_i^r(C_1^{i,r} + C_2^{i,r})}{4}\bigg) + \frac{45 \sigma^2}{4S} \sum_{i =1}^{N}\omega_ip_i^rC_1^{i,r}, \label{thm: descent5}
\end{align}where $D_1^r \triangleq \sum_{i=1}^{N} \frac{9}{p_i^r(C_1^{i,r} + C_2^{i,r})}$. Lastly, taking the telescoping sum over \eqref{thm: descent5} gives rise to
\begin{align}
	&\frac{1}{R}\sum_{r = 0}^{R - 1}\E[\|\nabla f(\xb_0^r)\|^2] \notag \\
	\leq &\frac{2D_1(P^0- \ul f)}{\beta R}+\frac{10D_1\sigma^2}{SR}\sum_{r = 0}^{R - 1} \sum_{i =1}^{N}\omega_i\bigg(1+ \frac{p_i^r(C_1^{i,r} + C_2^{i,r})}{4}\bigg)+ \frac{45 \sigma^2}{4SR} \sum_{r = 0}^{R - 1}\sum_{i =1}^{N}\omega_ip_i^rC_1^{i,r}\\
	= &\frac{2D_1(P^0- \ul f)}{\beta R} + \frac{5D_2\sigma^2}{4S} ,\label{thm: descent6}
\end{align}where $D_1 \triangleq \max\limits_{r}D_1^{r}$, and $D_2 \triangleq \frac{1}{R}\sum_{r = 0}^{R - 1}\sum_{i=1}^{N}\omega_i(D_1(8 + 2p_i^r(C_1^{i,r} + C_2^{i,r})) + 9p_i^rC_1^{i,r})$. This completes the proof.
\hfill $\blacksquare$

\section{Proof of Corollary 1}
\label{sec: cor1_proof}

\noindent {\bf Proof:} since $p_i^r  = \frac{m}{N}$ and $Q_i^r = Q$, we can set $ \eta_{i} = \eta, \gamma_i = \gamma, d_i^r = d, a_i^r = a$ and then we have 
\begin{align}
	&C_1^{i,r} = C_1 = (a + d)\gamma\eta \wt Q, C_2^{i,r} =C_2 =  a \gamma \eta \wt Q.
\end{align}
Following the same procedure as that in the proof of Theorem 1, we get
\begin{align}
	&p_i(C_1 + C_2)\sum_{i=1}^{N}\omega_i\Xi_i^r +\frac{1}{2\beta^2 }\E[\|\xb_0^{r+1} - \xb_0^r\|^2] 
	\leq  	\frac{2(P^r - P^{r+1} )}{\beta }+\frac{10 \sigma^2}{S} \bigg(1+ \frac{p_i(C_1+ C_2)}{4}\bigg),	\label{cor1: descent1} \\
	&	\E[\|\nabla f(\xb_{0}^r)\|^2] 
	\leq   \frac{1 + c}{\beta^2}\E[\|\xb_0^r - \xb_0^{r+1} \|^2]+ \frac{1 + 1/c}{2}(2-p_iC_1)\sum_{i =1}^{N}\omega_i\Xi_i^r + \frac{5(1 + 1/c) \sigma^2}{4S} p_iC_1.	\label{cor1: descent2}
\end{align}
Then, combing \eqref{cor1: descent1} and \eqref{cor1: descent2} yields
\begin{align}
	\E[\|\nabla f(\xb_{0}^r)\|^2] 
	\leq & 	\frac{2D_1(P^r - P^{r+1} )}{\beta }+\frac{10D_1 \sigma^2}{S}\bigg(1+ \frac{p_i(C_1 + C_2)}{4}\bigg) + (1 + \frac{1}{c})\frac{5 \sigma^2}{4S}p_iC_1,\label{cor1: descent3}
\end{align}
where 
\begin{align}
	D_1 \triangleq 2(1 + c) + \frac{1}{2}(1 + \frac{1}{c}) \frac{2-p_iC_1}{p_i(C_1 + C_2)} = (1 + \frac{1}{c})\frac{2N}{m(C_1 + C_2)}  + 2(1+c) - (1 + \frac{1}{c})\frac{C_1}{2(C_1 + C_2)}.
\end{align}
Let $c = \frac{1}{8}$, then we have $2(1+c) - (1 + \frac{1}{c})\frac{C_1}{2(C_1 + C_2)} \leq 0$, which implies that $D_1 \leq \frac{18N}{m(C_1 + C_2)}$. Substituting the results into the RHS of \eqref{cor1: descent3} gives rise to 
\begin{align}
	\E[\|\nabla f(\xb_{0}^r)\|^2] 
	\leq & 	\frac{36N(P^r - P^{r+1} )}{\beta m (C_1 + C_2) }+\frac{180N\sigma^2}{m(C_1 + C_2)S} + \frac{45 \sigma^2}{S} + \frac{45 m\sigma^2C_1}{4NS}. \label{cor1: descent4}
\end{align}
Lastly, taking the telescoping sum over \eqref{cor1: descent4} yields
\begin{align}
	\frac{1}{R}\sum_{r = 0}^{R - 1}\E[\|\nabla f(\xb_{0}^r)\|^2] 
	\leq & 	\frac{36N(P^0 - \ul f )}{\beta m (C_1 + C_2)R }+\frac{180N\sigma^2}{m(C_1 + C_2)S} + \frac{45 \sigma^2}{S} + \frac{45 m\sigma^2C_1}{4NS}. 
\end{align}
Therefore, FedVRA achieves a convergence rate $\mathcal{O}\bigg(\frac{N}{mR} + \frac{N\sigma^2}{m\sqrt{R}}\bigg)$ as $S =\sqrt{R}$. \hfill $\blacksquare$

\section{Proof of Theorem 2}
\label{sec: thm2_proof}

\subsection{Preliminary}
To establish the convergence property of FedVRA-U, let us make the following assumptions.

\vspace{-0.3cm}
\begin{assumption}[\textbf{Lower boundedness and L-smoothness}] \label{assumption: lower-bounded}
	All local cost functions $f_i(\cdot, \cdot)$ in problem (25) are non-convex and lower bounded, i.e., $f_i(\xb,\yb_i) \geq \underline{f} > {-\infty},~\forall \xb\in \Xc, \yb_i\in \Yc_i$, and $L_i$-smooth, which implies that $
	\|\nabla f_i(\xb, \yb_i) - \nabla f_i(\xb^\prime, \yb_i^\prime)\| \leq L_i \|(\xb, \yb_i)-(\xb^\prime, \yb_i^\prime)\|, \forall \xb\in \Xc, \yb_i\in \Yc_i$.
\end{assumption}

\vspace{-0.3cm}
\begin{assumption}[Bounded variance] \label{assumption: SGD_variance}
	For a data sample $\xi_i$ uniformly sampled at random from $\Dc_i$, both of the resulting stochastic gradients (SGD) for problem (25) are unbiased and have bounded variances, i.e.,
	\begin{align}
		& \E[\nabla f_i(\xb_i, \yb_i; \xi_i)] = \nabla f_i(\xb_i, \yb_i), \\
		&\E[\|\nabla f_i(\xb_i, \yb_i; \xi_i) - \nabla f_i(\xb_i, \yb_i)\|^2] \leq \sigma^2,
	\end{align}where $\sigma > 0$ is a constant.
\end{assumption}
Let  $g_i(\xb_i)$ denote the SG of $f_i(\cdot)$ at $\xb_i$ over a mini-batch of $S$ samples i.i.d drawn from $\Dc_i$, i.e., $g_i(\xb_i) \triangleq \frac{1}{S} \sum_{\xi_i} \nabla f_i(\xb_i; \xi_i)$. Then, $g_i(\xb_i)$ is also unbiased with variance bounded by $\frac{\sigma^2}{S}$.

Besides, we define the following term as the optimality gap between a stationary solution of problem (25) in the manuscript.
\begin{align}
	G^r &=  \frac{1}{\beta^2}\|\xb_0^r - \Pc_{\Xc}(\xb_0^r - \beta\nabla_x f(\xb_0^r, \yb^{r+1}))\|^2 +\sum_{i = 1}^{N}\frac{\omega_i }{(\eta_{i}^y)^2}\|\yb_i^r - \Pc_{\Yc_i}(\yb_i^r - \eta_{y_i}^r\nabla_y f_i(\xb_0^r, \yb_i^r))\|^2.
\end{align} 
It can be verified that $G^r = 0$ implies that $(\xb_0^r, \yb^{r+1})$ is a stationary point of problem (25) in the manuscript. We also define the virtual sequence $ \{(\wt \xb_i^{r}, \wt \yb_i^{r}, \wt \lambdab_i^{r})\}$ by assuming that all clients are active at round $r$, i.e., $\forall i$,
\begin{subequations}
	\begin{align}
		&\wt \xb_i^{r, 0} = \xb_0^{r},~ \wt \yb_i^{r, 0} = \yb_i^{r-1},\\
		&\wt \xb_i^{r, t+1} = \wt \xb_i^{r, t},~	\wt \yb_i^{r, t+1} = \Pc_{\Yc_i}\{\wt \yb_{i}^{r, t} - \eta_{i}^yg_i^{y}(\wt \yb_{i}^{r, t})\}, ~\wt \yb_i^{r+1} = \wt \yb_i^{i, Q_{y_i}^r}, 0 \leq t \leq Q_{y_i}^r - 1, \\
		& \wt \xb_i^{r, t+1} =\wt \xb_i^{r,t} - \eta_i(g_i^x(\wt \xb_i^{r,t}) - \lambdab_i^{r} + \gamma_i(\wt \xb_i^{r,t} - \xb_0^r)),\wt \xb_i^{r+1} = \wt \xb_i^{r, \hat Q_i^r}, ~\wt \yb_i^{r, t+1} = \wt \yb_i^{r, t},Q_{y_i}^r \leq t \leq \hat Q_i^r -1,\\
		&\wt \lambdab_{i}^{r+1} =  \lambdab_{i}^{r} +  a_i^r\gamma_i(\xb_{0}^r - \wt \xb_i^{r+1}).
	\end{align}
\end{subequations}Besides, we redefine the terms $C_1^{i,r}, C_2^{i,r}, \Xi_i^r$ that will be used in the proof.
\begin{align}
	&C_1^{i,r} \triangleq (a_i^r + d_i^r)\gamma_i\eta_i \wt Q_i^{r}, ~C_2^{i,r} \triangleq a_i^r\gamma_i\eta_i \wt Q_i^{r}, \\
	&\Xi_i^r \triangleq \E[\| \nabla_x f_i(\xb_0^r, \yb_{i}^r) - \lambdab_i^r\|^2], 
	~P^r \triangleq \E[f(\xb_0^r, \yb^r)] + \sum_{i = 1}^{N}\omega_i\frac{8\beta\Xi_i^r}{p_i^rC_2^{i,r}(1-p_i^r)}. \label{def: potential}
\end{align}

\subsection{Technical lemmas and their proofs}

\noindent \fbox{\parbox{1\linewidth}{ 
		\begin{Lemma} \label{lem: dual_opt}
			For any round $r$ and client $i$, it holds that
			\begin{align}
				& \sum_{t= Q_{y_i}^r}^{\hat Q_i^r - 1}\frac{b_i^{r,t}}{\|\bb_i^r\|_1}\E[\|\nabla f_i(\xb_0^r, \wt \yb_i^{r+1})  -g_i(\wt \xb_i^{r,t}, \wt \yb_i^{r+1})\|^2] 
				\leq \frac{1}{2}\Xi_i^r + \frac{5\sigma^2}{4S}.
			\end{align}
		\end{Lemma}
}}
\newline

\noindent {\bf Proof:} It can be obtained by following the same procedure as the proof of Lemma  \ref{lem: effect_dual}.
\hfill $\blacksquare$
\newline

\noindent \fbox{\parbox{1\linewidth}{ 
		\begin{Lemma} \label{lem: opt_descent_x}
			For any round $r$ and client $i$, it holds that 
			\begin{align}
				&\Xi_i^{r+1} - \Xi_i^r \notag \\
				\leq & \bigg(1 + \frac{4}{p_i^r C_2^{i, r}}\bigg)L_i^2 \E[\|\xb_0^{r+1}-\xb_0^r\|^2] - \bigg(\frac{(1-p_i^r)p_i^rC_2^{i,r}}{4} + \frac{(8+p_i^rC_2^{i,r} - 2p_i^r)(p_i^rC_2^{i,r})^2}{32}\bigg)\Xi_i^r \notag \\
				&+ \bigg(1+ \frac{p_i^r C_2^{i, r}}{4}\bigg)\bigg(1+ \frac{4}{p_i^r C_2^{i, r}}\bigg)p_i^r\bigg(1-\frac{1}{2}C_2^{i,r}\bigg)L_i^2\E[\|\wt \yb_i^{r+1} -  \yb_i^{r}\|^2] +\frac{5\sigma^2p_i^rC_2^{i,r}}{4S} \bigg(1+ \frac{p_i^r C_2^{i, r}}{4}\bigg). \label{lem: opt_descent_x_bd}
			\end{align}
\end{Lemma}}}
\newline

\noindent{\bf Proof:}
By the definition of $\Xi_i$, we have
\begin{align}
	&\Xi_i^{r+1} \notag \\
	=& \E[\|\nabla_x f_i(\xb_0^{r+1}, \yb_i^{r+1})-\nabla_x f_i(\xb_0^r, \yb_i^{r+1}) + \nabla_x f_i(\xb_0^r, \yb_i^{r+1}) - \lambdab_i^{r+1}\|^2]	\notag \\
	\leq & (1 + \frac{1}{\epsilon_i}) \E[\|\nabla_x f_i(\xb_0^{r+1}, \yb_i^{r+1})-\nabla_x f_i(\xb_0^r, \yb_i^{r+1}) \|^2] +(1 + \epsilon_i) \E[\| \nabla_x f_i(\xb_0^r, \yb_i^{r+1}) - \lambdab_i^{r+1}\|^2] \label{lem: opt2_bd1}\\
	= & (1 + \frac{1}{\epsilon_i})L_i^2 \E[\|\xb_0^{r+1}-\xb_0^r\|^2] +(1 + \epsilon_i)(1-p_i^r) \Xi_i^r +(1 + \epsilon_i)p_i^r\E[\| \nabla_x f_i(\xb_0^r, \wt \yb_i^{r+1}) - \wt \lambdab_i^{r+1}\|^2], \label{lem: opt2_bd2}
\end{align}where $\epsilon_i > 0$; \eqref{lem: opt2_bd1} follows by the fact that $(z_1 + z_2)^2 \leq (1 + \frac{1}{c}) z_1^2 + (1 + c)z_2^2, \forall c > 0$; \eqref{lem: opt2_bd2} holds because $\Prob(i \in \Ac^r) = p_i^r$. Then, we proceed to bound $\E[\| \nabla_x f_i(\xb_0^r, \wt \yb_i^{r+1}) - \wt \lambdab_i^{r+1}\|^2] $ by
\begin{align}
	&\E[\| \nabla_x f_i(\xb_0^r, \wt \yb_i^{r+1}) - \wt \lambdab_i^{r+1}\|^2] \notag \\
	=&\E\bigg[\bigg\| \nabla_x f_i(\xb_0^r, \wt \yb_i^{r+1}) - (1-C_2^{i,r}) \lambdab_i^{r}-C_2^{i,r}\sum_{t= Q_{y_i}^r}^{\hat Q_i^r - 1}\frac{b_i^{r,t}}{\|\bb_i^r\|_1}g_i(\wt \xb_i^{r,t}, \wt \yb_i^{r+1})\bigg\|^2\bigg] \label{lem: opt2_bd3} \\
	= & \E\bigg[\bigg\|(1-C_2^{i,r})( \nabla_x f_i(\xb_0^r, \wt \yb_i^{r+1}) - \lambdab_i^{r})+C_2^{i,r}\sum_{t= Q_{y_i}^r}^{\hat Q_i^r - 1}\frac{b_i^{r,t}}{\|\bb_i^r\|_1}( \nabla_x f_i(\xb_0^r,\wt \yb_i^{r+1})  -g_i(\wt \xb_i^{r,t}, \wt \yb_i^{r+1}))\bigg\|^2\bigg] \notag \\
	\leq & (1-C_2^{i,r})\E[\|\nabla_x f_i(\xb_0^r, \wt \yb_i^{r+1}) - \lambdab_i^{r}\|_2^2 +C_2^{i,r}\sum_{t= Q_{y_i}^r}^{\hat Q_i^r - 1}\frac{b_i^{r,t}}{\|\bb_i^r\|_1}\E[\|\nabla_x f_i(\xb_0^r, \wt \yb_i^{r+1}) -g_i(\wt \xb_i^{r,t}, \wt \yb_i^{r+1})\|^2]  \notag\\
	\leq &(1-C_2^{i,r}/2)\E[\| \nabla_x f_i(\xb_0^r, \wt \yb_i^{r+1}) - \lambdab_i^{r}\|^2]  + \frac{5\sigma^2}{4S}C_2^{i,r}, \label{lem: opt2_bd5}
\end{align}where  \eqref{lem: opt2_bd3} follows by the definition of $\wt \lambdab_i^{r+1}$; \eqref{lem: opt2_bd5} follows thanks to Lemma \ref{lem: dual_opt}. Next, substituting \eqref{lem: opt2_bd5} into \eqref{lem: opt2_bd2} yields
\begin{align}
	\Xi_i^{r+1} 
	\leq & (1 +1/\epsilon_i)L_i^2 \E[\|\xb_0^{r+1}-\xb_0^r\|^2] +(1 + \epsilon_i)(1-p_i^r) \Xi_i^r +\frac{5\sigma^2(1 + \epsilon_i)p_i^rC_2^{i,r}}{4S}\notag \\
	&+(1 + \epsilon_i)p_i^r(1-C_2^{i,r}/2)\E[\|| \nabla_x f_i(\xb_0^r, \wt \yb_i^{r+1}) - \lambdab_i^{r}\|^2] 
	.\label{lem: opt2_bd6}
\end{align}
Furthermore, note that
\begin{align}
	&\E[\|\nabla_x f_i(\xb_0^r, \wt \yb_i^{r+1}) - \lambdab_{i}^r\|^2] \notag \\
	=&	\E[\|\nabla_x f_i(\xb_0^r, \wt \yb_i^{r+1}) - \nabla_x f_i(\xb_0^r, \yb_i^{r})  +  \nabla_x f_i(\xb_0^r, \yb_i^{r}) - \lambdab_{i}^r\|^2] \notag \\
	\leq & (1+ \epsilon_i)\Xi_i^r + (1+ 1/\epsilon_i)\E[\|\nabla_x f_i(\xb_0^r, \wt \yb_i^{r+1}) - \nabla_x f_i(\xb_0^r, \yb_i^{r})\|^2] \label{lem: opt2_bd6_1}\\
	\leq &  (1+ \epsilon_i)\Xi_i^r + (1+ 1/\epsilon_i)L_i^2\E[\|\wt \yb_i^{r+1} -  \yb_i^{r}\|^2],\label{lem: opt2_bd6_2}
\end{align}where \eqref{lem: opt2_bd6_1} follows by the fact that $(z_1 + z_2)^2 \leq (1 + \frac{1}{c}) z_1^2 + (1 + c)z_2^2, \forall c > 0$;  \eqref{lem: opt2_bd6_2} holds due to Assumption 1. As a result, we have
\begin{align}
	\Xi_i^{r+1}
	\leq & (1 + 1/\epsilon_i)L_i^2 \E[\|\xb_0^{r+1}-\xb_0^r\|^2] +(1 + \epsilon_i)(1 + p_i^r\epsilon_i-(1+\epsilon_i)p_i^rC_2^{i,r}/2) \Xi_i^r \notag \\
	&+(1 + \epsilon_i)(1+1/\epsilon_i)p_i^r(1-C_2^{i,r}/2)L_i^2\E[\|\wt \yb_i^{r+1} -  \yb_i^{r}\|^2]
	+\frac{5\sigma^2(1 + \epsilon_i)p_i^rC_2^{i,r}}{4S}, 
\end{align}
which implies that
\begin{align}
	\Xi_i^{r+1} - \Xi_i^r
	\leq & (1 + 1/\epsilon_i)L_i^2 \E[\|\xb_0^{r+1}-\xb_0^r\|^2] + ((1+ \epsilon_i)(1 + p_i^r\epsilon_i - (1 + \epsilon_i)p_i^rC_2^{i,r}/2)-1)\Xi_i^r\notag \\
	&+ (1+ \epsilon_i)(1+ 1/\epsilon_i)p_i^r(1-C_2^{i,r}/2)L_i^2\E[\|\wt \yb_i^{r+1} -  \yb_i^{r}\|^2] +\frac{5\sigma^2(1 + \epsilon_i)p_i^rC_2^{i,r}}{4S}.\label{lem: opt2_bd7}
\end{align}
Let us pick $\epsilon_i = \frac{p_i^r C_2^{i, r}}{4}$, then it holds that
\begin{align}
	(1+ \epsilon_i)(1 + p_i^r\epsilon_i - (1 + \epsilon_i)p_i^rC_2^{i,r}/2)-1
	=  - \frac{(1-p_i^r)p_i^rC_2^{i,r}}{4} - \frac{(8+p_i^rC_2^{i,r} - 2p_i^r)(p_i^rC_2^{i,r})^2}{32},
\end{align}
and we obtain from \eqref{lem: opt2_bd7} that
\begin{align}
	&\Xi_i^{r+1} - \Xi_i^r\notag \\
	\leq & \bigg(1 + \frac{4}{p_i^r C_2^{i, r}}\bigg)L_i^2 \E[\|\xb_0^{r+1}-\xb_0^r\|^2] - \bigg(\frac{(1-p_i^r)p_i^rC_2^{i,r}}{4} + \frac{(8+p_i^rC_2^{i,r} - 2p_i^r)(p_i^rC_2^{i,r})^2}{32}\bigg)\Xi_i^r \notag \\
	&+ \bigg(1+ \frac{p_i^r C_2^{i, r}}{4}\bigg)\bigg(1+ \frac{4}{p_i^r C_2^{i, r}}\bigg)p_i^r\bigg(1-\frac{1}{2}C_2^{i,r}\bigg)L_i^2\E[\|\wt \yb_i^{r+1} -  \yb_i^{r}\|^2] +\frac{5\sigma^2p_i^rC_2^{i,r}}{4S} \bigg(1+ \frac{p_i^r C_2^{i, r}}{4}\bigg).
\end{align} This completes the proof.\hfill $\blacksquare$

\noindent \fbox{\parbox{1\linewidth}{ 
		\begin{Lemma} \label{lem: optimality_gap2}
			For any round $r$, it holds that
			\begin{align}
				\E[G^r] 
				\leq & 2\sum_{i=1}^{N}\omega_i(1+p_i^r- p_i^rC_1^{i,r})\E[\|\nabla_x f_i(\xb_0^r,  \yb_i^{r})-\lambdab_i^{r}\|^2] + \frac{2}{\beta^2}\E[\|\xb_0^r  - \xb_0^{r+1}\|^2] + \frac{\sigma^2}{2S}\sum_{i=1}^{N}\omega_i (5p_i^r + 8)\notag \\
				&+ \sum_{i=1}^{N}\omega_i\bigg(\frac{6Q_{y_i}^r - 4}{(\eta_{i}^{y})^2} + 2(4-C_1^{i,r})Q_{y_i}^rL_i^2 - 4L_i^2\bigg) \sum_{t = 0}^{Q_{y_i}^r - 1} \E[\| \wt \yb_i^{r, t+1} - \wt\yb_i^{r, t}\|_2^2]. \label{lem: optimality_gap2_bd}
			\end{align}
\end{Lemma}}}
\newline

\noindent {\bf Proof:} By the definition of $G^r$, we have
\begin{align}
	&\E[G^r] \notag \\
	=&  \frac{1}{\beta^2}\E[\|\xb_0^r - \Pc_{\Xc}(\xb_0^r - \beta\nabla_x f(\xb_0^r, \yb^{r+1}))\|^2] +\sum_{i = 1}^{N}\frac{\omega_i }{(\eta_{i}^{y})^2}\E[\|\yb_i^r - \Pc_{\Yc_i}(\yb_i^r - \eta_{i}^{y}\nabla_y f_i(\xb_0^r, \yb_i^r))\|^2]. \label{lem: gap_bd1}
\end{align}
We proceed to bound the RHS terms of \eqref{lem: gap_bd1}. For the first term, we obtain
\begin{align}
	&\E[\|\xb_0^r - \Pc_{\Xc}(\xb_0^r - \beta\nabla_x f(\xb_0^r, \yb^{r+1}))\|^2] \notag \\
	= &\E[\|\xb_0^r  - \xb_0^{r+1} + \Pc_{\Xc}(\ub^{r+1}) - \Pc_{\Xc}(\xb_0^r - \beta\nabla_x f(\xb_0^r, \yb^{r+1}))\|^2] \label{lem: gap_bd2}\\
	\leq & 2\E[\|\xb_0^r  - \xb_0^{r+1}\|^2] + 2 \E[\|\ub^{r+1} -\xb_0^r + \beta\nabla_x f(\xb_0^r, \yb^{r+1})\|^2] \label{lem: gap_bd3}
\end{align}
\begin{align}
	=& 2\beta^2\E[\|\nabla_x f(\xb_0^r, \yb^{r+1}) + \sum_{i = 1}^{N}\omega_id_i^r\gamma_i (\xb_i^{r+1} - \xb_0^r) - \sum_{i = 1}^{N}\omega_i\lambdab_i^{r+1}\|_2^2]+2\E[\|\xb_0^r  - \xb_0^{r+1}\|^2] \label{lem: gap_bd4}\\
	\leq& 2\beta^2\sum_{i=1}^{N}\omega_i\E[\|\nabla_x f_i(\xb_0^r, \yb_i^{r+1}) + d_i^r\gamma_i (\xb_i^{r+1} - \xb_0^r) - \lambdab_i^{r+1}\|^2] +2\E[\|\xb_0^r  - \xb_0^{r+1}\|^2]\label{lem: gap_bd5} \\
	\leq & 2\beta^2\sum_{i=1}^{N}\omega_i	(1+p_i^r- p_i^rC_1^{i,r})\Xi_i^r + 2\beta^2\sum_{i=1}^{N}\omega_ip_i^r(2-C_1^{i,r})L_i^2\E[\| \wt \yb_i^{r+1}-\yb_i^{r}\|^2] \notag \\
	& +  \frac{5\beta^2\sigma^2}{2S}\sum_{i=1}^{N}\omega_i p_i^r C_1^{i,r}+ 2\E[\|\xb_0^r  - \xb_0^{r+1}\|^2],\label{lem: gap_bd6}
\end{align}
where \eqref{lem: gap_bd2} follows by the definition of $\ub^{r+1}$;  \eqref{lem: gap_bd3} follows due to the Jensen's Inequality and the non-expensiveness of projection operator; 
\eqref{lem: gap_bd4} holds by \eqref{lem: obj_descent_bd2}; \eqref{lem: gap_bd5} holds by the convexity of $\|\cdot\|^2$; \eqref{lem: gap_bd6} follows because of \eqref{lem: obj_descent_bd12}. For the second term, we have 
\begin{align}
	&\E[\|\yb_i^r - \Pc_{\Yc_i}(\yb_i^r - \eta_{i}^{y}\nabla_y f_i(\xb_0^r, \yb_i^r))\|^2] \notag \\
	= & \E[\|\yb_i^r - \wt \yb_i^{r+1}+\Pc_{\Yc_i}(\wt \yb_i^{r, Q_{y_i}^r - 1} - \eta_{i}^{y}\vb_i^{r, Q_{y_i}^r - 1})- \Pc_{\Yc_i}(\yb_i^r - \eta_{i}^{y}\nabla_y f_i(\xb_0^r, \yb_i^r))\|^2] \label{lem: gap_bd7}\\
	\leq & 2 \E[\|\yb_i^r - \wt \yb_i^{r+1}\|^2] + 4 \E[\|\wt \yb_i^{r, Q_{y_i}^r - 1}-\yb_i^r\|^2]+4(\eta_{i}^{y})^2\E[\|\vb_i^{r, Q_{y_i}^r - 1}-\nabla_y f_i(\xb_0^r, \yb_i^r)\|^2] \label{lem: gap_bd8}\\
	\leq & 2 \E[\|\yb_i^r - \wt \yb_i^{r+1}\|_2^2] + 4 \E[\|\wt \yb_i^{r, Q_{y_i}^r - 1}-\yb_i^r\|_2^2]+4 (\eta_{i}^{y})^2\E[\|\nabla_y f_i(\xb_0^r, \wt \yb_i^{r, Q_{y_i}^r - 1})-\nabla_y f_i(\xb_0^r, \yb_i^r)\|^2] \notag \\
	&+  \frac{4(\eta_{i}^{y})^2\sigma^2}{S} \label{lem: gap_bd9}\\
	\leq & 2 \E[\|\yb_i^r - \wt \yb_i^{r+1}\|^2] + 4 \E[\|\wt \yb_i^{r, Q_{y_i}^r - 1}-\yb_i^r\|^2]+4 (\eta_{i}^{y})^2L_i^2\E[\| \wt \yb_i^{r, Q_{y_i}^r-1} -\yb_i^r\|^2] +  \frac{4(\eta_{i}^{y})^2\sigma^2}{S}  \label{lem: gap_bd10}\\
	\leq & (6Q_{y_i}^r - 4 + 4(\eta_{i}^{y})^2L_i^2(Q_{y_i}^r-1)) \sum_{t = 0}^{Q_{y_i}^r - 1} \E[\| \wt \yb_i^{r, t+1} - \wt\yb_i^{r, t}\|^2] + \frac{4(\eta_{i}^y)^2\sigma^2}{S}, \label{lem: gap_bd11}
\end{align}
where \eqref{lem: gap_bd7} follows by the definition of $\wt \yb_i^{r+1}$; \eqref{lem: gap_bd8} follows by the Jensen's Inequality and the non-expensiveness of projection operator; \eqref{lem: gap_bd9} follows thanks to \cite[Lemma 4]{SCAFFOLD_2020}; \eqref{lem: gap_bd10} holds by Assumption 1; \eqref{lem: gap_bd11} holds by the Jensen's Inequality. Lastly, substituting the results of \eqref{lem: gap_bd6} and \eqref{lem: gap_bd11} into \eqref{lem: gap_bd1} gives rise to
\begin{align}
	&	\E[G^r] \notag \\
	\leq&2\sum_{i=1}^{N}\omega_i(1+p_i^r- p_i^rC_1^{i,r})\E[\|\nabla_x f_i(\xb_0^r,  \yb_i^{r})-\lambdab_i^{r}\|^2] + 2\sum_{i=1}^{N}\omega_ip_i^r(2-C_1^{i,r})L_i^2\E[\| \wt \yb_i^{r+1}-\yb_i^{r}\|^2] \notag \\
	& + \frac{2}{\beta^2}\E[\|\xb_0^r  - \xb_0^{r+1}\|^2] + \sum_{i=1}^{N}\frac{\omega_i}{(\eta_{i}^{y})^2}(6Q_{y_i}^r - 4 + 4(\eta_{i}^{y})^2L_i^2(Q_{y_i}^r-1)) \sum_{t = 0}^{Q_{y_i}^r - 1} \E[\| \wt \yb_i^{r, t+1} - \wt\yb_i^{r, t}\|_2^2] \notag \\
	&+  \frac{\sigma^2}{2S}\sum_{i=1}^{N}\omega_i (5p_i^r + 8)\notag \\
	\leq & 2\sum_{i=1}^{N}\omega_i(1+p_i^r- p_i^rC_1^{i,r})\E[\|\nabla_x f_i(\xb_0^r,  \yb_i^{r})-\lambdab_i^{r}\|^2] + \frac{2}{\beta^2}\E[\|\xb_0^r  - \xb_0^{r+1}\|^2] + \frac{\sigma^2}{2S}\sum_{i=1}^{N}\omega_i (5p_i^r + 8)\notag \\
	&+ \sum_{i=1}^{N}\omega_i\bigg(\frac{6Q_{y_i}^r - 4}{(\eta_{i}^{y})^2} + 2(4-C_1^{i,r})Q_{y_i}^rL_i^2 - 4L_i^2\bigg) \sum_{t = 0}^{Q_{y_i}^r - 1} \E[\| \wt \yb_i^{r, t+1} - \wt\yb_i^{r, t}\|_2^2].\label{lem: gap_bd12}
\end{align} This completes the proof.
\hfill $\blacksquare$

\noindent \fbox{\parbox{1\linewidth}{ 
		\begin{Lemma} \label{lem: obj_descent_xy}
			For any round $r$, it holds that
			\begin{align}
				& \E[f(\xb_0^{r+1},\yb^{r+1}) - f(\xb_0^r, \yb^{r+1})] \notag \\
				\leq&	\beta \sum_{i = 1}^{N}\omega_i(1+p_i^r- p_i^rC_1^{i,r})\E[\|\nabla_x f_i(\xb_0^r,  \yb_i^{r})-\lambdab_i^{r}\|^2]+ \beta \sum_{i = 1}^{N}\omega_ip_i^r(2 - C_1^{i,r})L_i^2\E[\| \wt \yb_i^{r+1}-\yb_i^{r}\|^2]\notag \\
				& -\bigg(\frac{3}{4\beta} -\sum_{i = 1}^{N} \omega_i  \frac{L_{i}}{2}\bigg)\E[\|\xb_0^{r+1} - \xb_0^r\|^2]  -\sum_{i = 1}^{N}\omega_ip_i^r\bigg(\frac{1}{\eta_{i}^{y}} - \frac{L_i}{2}\bigg)\sum_{t = 0 }^{Q_{y_i}^r - 1}
				\E[\|\wt \yb_i^{r, t+1} - \wt \yb_i^{r, t}\|_2^2] \notag \\
				&+ \frac{5\beta\sigma^2}{4S}\sum_{i=1}^{N}\omega_ip_i^rC_1^{i,r}. \label{lem: obj_descent_xy_bd}
			\end{align}
\end{Lemma}}}
\newline

\noindent {\bf Proof:} The poof is similar to that of Theorem 1 and it is also derived by analyzing the one-round progress of $f(\xb_{0}, \yb)$ with respect to $(\xb_{0}, \yb)$. Compared with Theorem 1, we need to specifically consider the effect of the constraints $\Xc$ and $\Yc_i$ and the update of $\yb_{i}$ to the construction of the potential function. We follow the similar strategy in \cite{NESTT_2016} to tackle this issue.

\noindent \underline{\bf One round analysis w.r.t. $\yb$:}
According to \cite[Lemma 3.2]{PALM_2014}, the update of $\yb_i$ implies that $\forall i \in \Ac^r, \forall 0 \leq t \leq Q_{y_i}^r - 1$,
	\begin{align} 
		&\E[f_i(\xb_0^{r}, \yb_i^{r, t+1}) - f_i(\xb_0^{r}, \yb_i^{r, t})] 	\leq -\bigg(\frac{1}{\eta_{i}^{y}} - \frac{L_{i}}{2}\bigg)\E[\|\yb_i^{r, t+1} - \yb_i^{r, t}\|_2^2], \label{thm2: descent_yi}
	\end{align}Summing up \eqref{thm2: descent_yi} from $t = 0$ to $Q_{y_i}^r - 1$ yields
	\begin{align}
		\E[f_i(\xb_0^{r}, \yb_i^{r})] - \E[f_i(\xb_0^{r}, \yb_i^{r+1})] 
		=&\E[f_i(\xb_i^{r, 0}, \yb_i^{r, 0})] - \E[f_i(\xb_i^{r, Q_{y_i}^r}, \yb_i^{r, Q_{y_i}^r})] \label{thm2: obj_descent_yi1} \\ 
		= & \sum_{t = 0 }^{Q_{y_i}^r - 1}\E[f_i(\xb_i^{r, t+1}, \yb_i^{r, t+1}) - f_i(\xb_i^{r, t}, \yb_i^{r, t})] \notag \\
		\leq& -\bigg(\frac{1}{\eta_{i}^{y}} - \frac{L_i}{2}\bigg)\sum_{t = 0 }^{Q_{y_i}^r - 1}
		\E[\|\yb_i^{r, t+1} - \yb_i^{r, t}\|_2^2].\label{thm2: obj_descent_yi2}
	\end{align}where \eqref{thm2: obj_descent_yi1} follows by the fact that $\xb_i^{r, 0} = \xb_i^{r, Q_{y_i}^r} = \xb_0^r, \yb_i^{r, Q_{y_i}^r} = \yb_i^{r+1}$. As a result, the objective function $f$ descends with local updates of $\yb$ as follows
	\begin{align}
		\E[f(\xb_0^r, \yb^r) - f(\xb_0^r, \yb^{r+1})] 
		= &\E\bigg[ \sum_{i \in \Ac^r}\omega_i\bigg(f_i(\xb_0^{r}, \yb_i^{r, 0}) - f_i(\xb_0^{r}, \yb_i^{r+1})\bigg)\bigg] \notag \\ 
		= &\sum_{i =1}^{N}\omega_ip_i^r\E[ f_i(\xb_0^{r}, \yb_i^{r, 0}) - f_i(\xb_0^{r}, \wt \yb_i^{r+1})] \notag \\ 
		\leq& -\sum_{i = 1}^{N}\omega_ip_i^r\bigg(\frac{1}{\eta_{i}^{y}} - \frac{L_i}{2}\bigg)\sum_{t = 0 }^{Q_{y_i}^r - 1}
		\E[\|\wt \yb_i^{r, t+1} - \wt \yb_i^{r, t}\|_2^2],
		\label{thm2: descent_y}
	\end{align}where \eqref{thm2: descent_y} follows by the fact that $\xb_i^{r, t} = \xb_0^r, \forall 0 \leq t \leq Q_{y_i}^r$, and $\yb_i^{r, Q_{y_i}^r} = \yb_i^{r+1}$.

\noindent \underline{\bf One round analysis w.r.t. $\xb_0$:} According to Assumption 1, we have
\begin{align}
	& \E[f(\xb_0^{r+1},\yb^{r+1}) - f(\xb_0^r, \yb^{r+1})] \notag \\
	=& \E\bigg[\sum_{i = 1}^{N} \omega_i f_i(\xb_{0}^{r+1}, \yb_{i}^{r+1}) - 	\sum_{i = 1}^{N} \omega_i f_i(\xb_{0}^{r}, \yb_i^{r+1}) \bigg]\notag \\
	\leq & \sum_{i = 1}^{N} \omega_i\E[ \langle \nabla_x f_i(\xb_0^r, \yb_i^{r+1}), \xb_0^{r+1} - \xb_0^r\rangle] + \sum_{i = 1}^{N} \omega_i \frac{L_{i}}{2} \E[\|\xb_0^{r+1} - \xb_0^r\|^2] \notag \\
	= & \E\bigg[\bigg\langle\sum_{i = 1}^{N}\omega_i\nabla_x f_i(\xb_0^r, \yb_i^{r+1}) + \frac{1}{\beta}(\xb_0^{r+1} - \xb_0^r) , \xb_0^{r+1} - \xb_0^r \bigg\rangle\bigg] \notag \\
	&-\sum_{i = 1}^{N} \omega_i \bigg(\frac{1}{\beta} - \frac{L_{i}}{2} \bigg)\E[\|\xb_0^{r+1} - \xb_0^r\|^2]. \label{lem: obj_descent_bd1}
\end{align}
Let us define $\ub^{r+1} \triangleq \xb_0^r + \beta\sum_{i = 1}^{N}\omega_id_i^r\gamma_i (\xb_i^{r+1} - \xb_0^r) - \beta\sum_{i = 1}^{N}\omega_i\lambdab_i^{r+1}$. By the update of $\xb_0^{r+1}$, we have
\begin{align}
	\xb_0^{r+1} = \arg\min_{\xb_0} \frac{1}{2\beta} \|\xb_0 - \ub^{r+1}\|_2^2 + \delta_{\Xc}(\xb_0). \label{lem5: bd2}
\end{align}
where $\delta_{\Xc}(\xb_0)$ is an indicator function of $\xb_0$ on the convex set $\Xc$. 
Then, the first-order optimality condition of problem \eqref{lem5: bd2} gives rise to $\xb_0^{r+1} - \ub^{r+1} + \beta \xi^{r+1} = 0$, where $\xi^{r+1} \in \partial \delta_{\Xc}(\xb_0^{r+1})$. As a result, we get $
\xb_0^{r+1} -\xb_0^r = \beta\bigg(\sum_{i = 1}^{N}\omega_id_i^r\gamma_i (\xb_i^{r+1} - \xb_0^r) -  \sum_{i = 1}^{N}\omega_i\lambdab_i^{r+1} - \xi^{r+1}\bigg)$.
Substituting the result into \eqref{lem: obj_descent_bd1} yields
\begin{align}
	&  \E[f(\xb_0^{r+1},\yb^{r+1}) - f(\xb_0^r, \yb^{r+1})]\notag \\
	\leq & \E\bigg[\bigg\langle\sum_{i = 1}^{N}\omega_i\nabla_x f_i(\xb_0^r, \yb_i^{r+1}) + \frac{1}{\beta}(\xb_0^{r+1} - \xb_0^r) + \xi^{r+1} , \xb_0^{r+1} - \xb_0^r \bigg\rangle\bigg] \notag \\
	&-\sum_{i = 1}^{N} \omega_i \bigg(\frac{1}{\beta} - \frac{L_{i}}{2} \bigg)\E[\|\xb_0^{r+1} - \xb_0^r\|^2]  \label{lem: obj_descent_bd3}\\
	\leq &  \beta \E\bigg[\bigg\|\sum_{i = 1}^{N}\omega_i\nabla_x f_i(\xb_0^r, \yb_i^{r+1}) + \frac{1}{\beta}(\xb_0^{r+1} - \xb_0^r) + \xi^{r+1} \bigg\|^2\bigg]\notag \\
	&-\bigg(\frac{3}{4\beta} -\sum_{i = 1}^{N} \omega_i  \frac{L_{i}}{2} \bigg)\E[\|\xb_0^{r+1} - \xb_0^r\|^2]  \label{lem: obj_descent_bd4}\\
	= &\beta \E\bigg[\bigg\|\sum_{i = 1}^{N}\omega_i\nabla_x f_i(\xb_0^r, \yb_i^{r+1}) + \sum_{i = 1}^{N}\omega_id_i^r\gamma_i(\xb_i^{r+1} - \xb_0^r) - \sum_{i = 1}^{N}\omega_i\lambdab_i^{r+1}\bigg\|^2\bigg]\notag \\
	&-\bigg(\frac{3}{4\beta} -\sum_{i = 1}^{N} \omega_i  \frac{L_{i}}{2} \bigg)\E[\|\xb_0^{r+1} - \xb_0^r\|^2]  \label{lem: obj_descent_bd5}\\
	\leq& \beta \sum_{i = 1}^{N}\omega_i\E[\|\nabla_x f_i(\xb_0^r, \yb_i^{r+1}) +d_i^r\gamma_i(\xb_i^{r+1} - \xb_0^r) - \lambdab_i^{r+1}\|^2] \notag \\
	&-\bigg(\frac{3}{4\beta} -\sum_{i = 1}^{N} \omega_i  \frac{L_{i}}{2}\bigg)\E[\|\xb_0^{r+1} - \xb_0^r\|^2],
	\label{lem: obj_descent_bd6}
\end{align}
where  \eqref{lem: obj_descent_bd3} follows by the convexity of $\delta_{\Xc}(\xb_0)$ and the fact that $\xi^{r+1} \in \partial \delta_{\Xc}(\xb_0^{r+1})$, i.e. $	\langle \xi^{r+1}, \xb_0^{r+1} - \xb_0^r \rangle \geq 0$;  \eqref{lem: obj_descent_bd4} follows by the Jensen's Inequality.
We proceed to bound the RHS terms of \eqref{lem: obj_descent_bd6}. Firstly, we have
\begin{align}
	&\E[\|\nabla_x f_i(\xb_0^r, \yb_i^{r+1}) +d_i^r\gamma_i(\xb_i^{r+1} - \xb_0^r) - \lambdab_i^{r+1}\|^2] \notag \\
	= &  (1-p_i^r)\Xi_i^r+p_i^r\E[\|\nabla_x f_i(\xb_0^r, \wt \yb_i^{r+1})+d_i^r\gamma_i(\wt\xb_i^{r+1} - \xb_0^r)-\wt\lambdab_i^{r+1}\|^2] \label{lem: obj_descent_bd7}\\
	= &(1-p_i^r)\Xi_i^r+p_i^r\E[\|\nabla_x f_i(\xb_0^r, \wt \yb_i^{r+1})+(a_i^r + d_i^r)\gamma_i(\wt\xb_i^{r+1} - \xb_0^r)-\lambdab_i^{r}\|^2] \label{lem: obj_descent_bd8}\\
	\leq &(1-p_i^r)\Xi_i^r +p_i^r\E\bigg[\bigg\|\nabla_x f_i(\xb_0^r, \wt \yb_i^{r+1})-\lambdab_i^{r} -C_1^{i,r}  \sum_{t= Q_{y_i}^r}^{\hat Q_i^r - 1}\frac{b_i^{r,t}}{\|\bb_i^r\|_1} (g_i(\wt \xb_i^{r,t}, \wt \yb_i^{r+1}) - \lambdab_i^r)\bigg\|^2\bigg] \label{lem: obj_descent_bd9}\\
	\leq & (1-p_i^r)\Xi_i^r + p_i^r(1-C_1^{i,r})\E[\|\nabla_x f_i(\xb_0^r, \wt \yb_i^{r+1})-\lambdab_i^{r}\|^2]  \notag \\
	&+ p_i^rC_1^{i, r}\sum_{t= Q_{y_i}^r}^{\hat Q_i^r - 1}\frac{b_i^{r,t}}{\|\bb_i^r\|_1}\E[\|\nabla_x f_i(\xb_0^r, \wt \yb_i^{r+1})-g_i(\wt \xb_i^{r,t}, \wt \yb_i^{r+1})\|^2], \label{lem: obj_descent_bd10}
\end{align}
where \eqref{lem: obj_descent_bd7} follows because $\Prob(i \in \Ac^r) = p_i^r$; \eqref{lem: obj_descent_bd8} follows by the definition of $\wt \lambdab_i^{r+1}$; \eqref{lem: obj_descent_bd10} follows holds due to the convexity of $\|\cdot\|^2$ and $C_1^{i,r} \leq 1$. By applying Lemma \ref{lem: dual_opt}, we have from  \eqref{lem: obj_descent_bd10} that
\begin{align}
	&\E[\|\nabla_x f_i(\xb_0^r, \yb_i^{r+1}) +\gamma_i(\xb_i^{r+1} - \xb_0^r) - \lambdab_i^{r+1}\|^2] \notag \\
	\leq & (1-p_i^r)\Xi_i^r + p_i^r(1-\frac{1}{2}C_1^{i,r})\E[\|\nabla_x f_i(\xb_0^r, \wt \yb_i^{r+1})-\lambdab_i^{r}\|^2]  + 
	\frac{5\sigma^2}{4S}p_i^r C_1^{i,r}\notag \\
	\leq & (1-p_i^r)\Xi_i^r + p_i^r(2-C_1^{i,r})\E[\|\nabla_x f_i(\xb_0^r,  \yb_i^{r})-\lambdab_i^{r}\|^2]  \notag \\
	&+ p_i^r(2-C_1^{i,r})\E[\|\nabla_x f_i(\xb_0^r, \wt \yb_i^{r+1})-\nabla_x f_i(\xb_0^r, \yb_i^{r})\|^2]+
	\frac{5\sigma^2}{4S}p_i^rC_1^{i,r} \notag \\	
	\leq & (1+p_i^r- p_i^rC_1^{i,r})\Xi_i^r + p_i^r(2 - C_1^{i,r})L_i^2\E[\| \wt \yb_i^{r+1}-\yb_i^{r}\|^2] + \frac{5\sigma^2}{4S}p_i^rC_1^{i,r},\label{lem: obj_descent_bd12}
\end{align}
where \eqref{lem: obj_descent_bd12} holds by Assumption 1. Substituting the result of \eqref{lem: obj_descent_bd12} into  \eqref{lem: obj_descent_bd6} yields
\begin{align}
	& \E[f(\xb_0^{r+1},\yb^{r+1}) - f(\xb_0^r, \yb^{r+1})] \notag \\
	\leq&	\beta \sum_{i = 1}^{N}\omega_i(1+p_i^r- p_i^rC_1^{i,r})\Xi_i^r+ \beta \sum_{i = 1}^{N}\omega_ip_i^r(2 - C_1^{i,r})L_i^2\E[\| \wt \yb_i^{r+1}-\yb_i^{r}\|^2]\notag \\
	& -\bigg(\frac{3}{4\beta} -\sum_{i = 1}^{N} \omega_i  \frac{L_{i}}{2}\bigg)\E[\|\xb_0^{r+1} - \xb_0^r\|^2] + \frac{5\beta\sigma^2}{4S}\sum_{i=1}^{N}\omega_ip_i^rC_1^{i,r}.\label{lem: obj_descent_bd13}
\end{align}
\noindent\underline{\bf Derivation of the main result:}  Combining the results of \eqref{thm2: descent_y} and \eqref{lem: obj_descent_bd13} gives rise to 
\begin{align}
	& \E[f(\xb_0^{r+1},\yb^{r+1}) - f(\xb_0^r, \yb^{r})] \notag \\
	\leq & \beta \sum_{i = 1}^{N}\omega_i(1+p_i^r- p_i^rC_1^{i,r})\E[\|\nabla_x f_i(\xb_0^r,  \yb_i^{r})-\lambdab_i^{r}\|^2]-\bigg(\frac{3}{4\beta} -\sum_{i = 1}^{N} \omega_i  \frac{L_{i}}{2}\bigg)\E[\|\xb_0^{r+1} - \xb_0^r\|^2]\notag \\
	&-\sum_{i = 1}^{N}\omega_ip_i^r\bigg(\frac{1}{\eta_{i}^{y}} - \frac{L_i}{2} - \beta (2 - C_1^{i,r})Q_{y_i}^rL_i^2\bigg)\sum_{t = 0 }^{Q_{y_i}^r - 1}
	\E[\|\wt \yb_i^{r, t+1} - \wt \yb_i^{r, t}\|_2^2] + \frac{5\beta\sigma^2}{4S}\sum_{i=1}^{N}\omega_ip_i^rC_1^{i,r}. \label{thm2: descent_xy_bd}
\end{align}
This completes the proof. \hfill $\blacksquare$

\subsection{Main proof of Theorem 2}

\noindent{\bf Proof:} By multiplying the two sides of \eqref{lem: opt_descent_x_bd} by $\frac{8\beta}{p_i^rC_2^{i,r}(1-p_i^r)}$ in Lemma \ref{lem: opt_descent_x}, and adding it to  \eqref{thm2: descent_xy_bd}, we have
\begin{align}
	&P^{r+1} - P^r \notag \\
	\leq &- \beta\sum_{i=1}^{N}\omega_i\bigg(p_i^rC_1^{i,r}+\frac{(8+p_i^rC_2^{i,r} - 2p_i^r)p_i^rC_2^{i,r}}{4(1-p_i^r)}\bigg)\E[\|\nabla_x f_i(\xb_0^r,  \yb_i^{r})-\lambdab_i^{r}\|_2^2] \notag \\
	&- \sum_{i = 1}^{N}\omega_ip_i^r\bigg(\frac{1}{\eta_{i}^{y}} - \frac{L_i}{2} - \beta (2 - C_1^{i,r})Q_{y_i}^rL_i^2- \frac{\beta(4+ p_i^r C_2^{i, r})^2(2-C_2^{i,r})Q_{y_i}^rL_i^2}{(p_i^rC_2^{i,r})^2(1-p_i^r)}\bigg)\sum_{t = 0 }^{Q_{y_i}^r - 1}
	\E[\|\wt \yb_i^{r, t+1} - \wt \yb_i^{r, t}\|_2^2] \notag \\
	&-\sum_{i = 1}^{N} \omega_i \bigg(\frac{3}{4\beta} - \frac{L_{i}}{2}-\frac{8\beta}{p_i^rC_2^{i,r}(1-p_i^r)}\bigg(1 + \frac{4}{p_i^r C_2^{i, r}}\bigg)L_i^2 \bigg)\E[\|\xb_0^{r+1} - \xb_0^r\|_2^2]\notag \\
	&+\frac{5\beta\sigma^2}{4S}\sum_{i=1}^{N}\omega_i\bigg(p_i^rC_1^{i,r}+ \frac{2(4 + p_i^r C_2^{i, r})}{1-p_i^r}\bigg),	\label{thm2: descent_xy_bd2}
\end{align}
Then, we claim that the coefficients for the second and the last term in the RHS of \eqref{thm2: descent_xy_bd2} satisfy
\begin{align}
	&\frac{3}{4\beta} - \frac{L_{i}}{2}-\frac{8\beta}{p_i^rC_2^{i,r}(1-p_i^r)}\bigg(1 + \frac{4}{p_i^r C_2^{i, r}}\bigg)L_i^2 \geq \frac{1}{4\beta}, \label{condition1}\\
	&\frac{1}{\eta_{i}^{y}} - \frac{L_i}{2} - \beta (2 - C_1^{i,r})Q_{y_i}^rL_i^2- \frac{4\beta(4+ p_i^r C_2^{i, r})^2(2-C_2^{i,r})Q_{y_i}^rL_i^2}{(p_i^rC_2^{i,r})^2(1-p_i^r)} \geq \frac{1}{2\eta_{i}^{y}} .\label{condition2}
\end{align}
To prove \eqref{condition1}, it suffices to show that $j = \arg\min_i \gamma_i,
1-  \frac{L_j}{\gamma_j} - \frac{16L_j^2}{\gamma_j^2p_j^rC_2^{j,r}(1-p_j^r)}\bigg(1 + \frac{4}{p_j^rC_2^{j,r}}\bigg)   \geq  0$, which holds if $
1-  \frac{L_j}{\gamma_j} - \frac{80L_j^2}{\gamma_j^2(p_j^rC_2^{j,r})^2(1-p_j^r)}  \geq  0$. By finding the root of the above quadratic inequality, we need
\begin{align}
	\gamma_j \geq \frac{L_j}{2} + \frac{L_j}{2}\sqrt{1+ \frac{320}{(p_j^rC_2^{j,r})^2(1-p_j^r)}},\label{condition1_3}
\end{align} which certainly holds as $\gamma_i \geq \frac{L_i}{2} + \frac{9L_i}{p_i^rC_2^{i,r}\sqrt{1-p_i^r}}, \forall i, r$. As for \eqref{condition2}, it suffices to show that
\begin{align}
	\frac{1}{2\eta_{i}^{y}} - \frac{L_i}{2} - 2\beta Q_{y_i}^rL_i^2- \frac{8\beta(4+ p_i^r C_2^{i, r})^2Q_{y_i}^rL_i^2}{(p_i^r C_2^{i, r})^2(1-p_i^r)} \geq 	0,\label{condition2_1}
\end{align}as $C_1^{i,r} \leq 1$ and $C_2^{i,r} \leq 1$. It can be checked that \eqref{condition2_1} is true as $\frac{1}{\eta_{i}^{y}} \geq L_i + 4\beta Q_{y_i}^rL_i^2 (1 + \frac{400}{(a_i^r\gamma_i\eta_i \wt Q_i^{r})^2})$ in Theorem 2. Therefore, we have from \eqref{thm2: descent_xy_bd2} that
\begin{align}
	&P^{r+1} -P^r \notag \\
	\leq &- \beta\sum_{i=1}^{N}\omega_i\bigg(p_i^rC_1^{i,r}+\frac{(8+p_i^rC_2^{i,r} - 2p_i^r)p_i^rC_2^{i,r}}{4(1-p_i^r)}\bigg)\E[\|\nabla_x f_i(\xb_0^r,  \yb_i^{r})-\lambdab_i^{r}\|_2^2] \notag \\
	&- \sum_{i = 1}^{N}\frac{\omega_ip_i^r}{2\eta_{i}^y}\sum_{t = 0 }^{Q_{y_i}^r - 1}
	\E[\|\wt \yb_i^{r, t+1} - \wt \yb_i^{r, t}\|_2^2] -\sum_{i = 1}^{N} \frac{\omega_i \gamma_i}{2}\E[\|\xb_0^{r+1} - \xb_0^r\|_2^2]\notag \\
	&+\frac{5\beta\sigma^2}{4S}\sum_{i=1}^{N}\omega_i\bigg(p_i^rC_1^{i,r}+ \frac{2(4 + p_i^r C_2^{i, r})}{1-p_i^r}\bigg).	\label{thm2: descent_xy_bd3}
\end{align}
Next, we combine \eqref{thm2: descent_xy_bd3} and  \eqref{lem: optimality_gap2_bd} in Lemma \ref{lem: optimality_gap2} and obtain
\begin{align}
	\E[G^r] \leq & D_3^r \bigg(\wt P^r - \wt P^{r+1}  + \frac{5\beta\sigma^2}{4S}\sum_{i=1}^{N}\omega_i\bigg(p_i^rC_1^{i,r}+ \frac{2(4 + p_i^r C_2^{i, r})}{1-p_i^r}\bigg)\bigg)  +\frac{\sigma^2}{2S}\sum_{i=1}^{N}\omega_i (5p_i^r + 8) \notag \\
	= & D_3^r(Q^r - Q^{r+1}) + \frac{D_4^r\sigma^2}{S}, \label{thm2: bd8}
\end{align}
where 
\begin{align}
	D_3^r \triangleq& \sum_{i = 1}^{N} \bigg( \frac{2(1-p_i^r)(1+p_i^r- p_i^rC_1^{i,r})}{\beta(4p_i^rC_1^{i,r}(1-p_i^r) + (8 +p_i^r C_2^{i, r}-2p_i^r)p_i^rC_2^{i,r})} +  \frac{4}{\beta^2\gamma_i}+ \frac{4(3Q_{y_i}^r - 2)}{p_i^r\eta_{i}^{y}} \notag \\
	&~~~~~~~~+ \frac{4\eta_{i}^{y}((4-C_1^{i,r})Q_{y_i}^rL_i^2 - 2L_i^2)}{p_i^r}\bigg), \\
	D_4^r \triangleq& \frac{5\sigma^2}{4S} \sum_{i=1}^{N} \omega_i\bigg(\beta D_3^r\bigg(p_i^rC_1^{i,r}+ \frac{2(4 + p_i^r C_2^{i, r})}{1-p_i^r}\bigg) + 2p_i^r + \frac{16}{5}\bigg).
\end{align}
Lastly, summing up the two sides of \eqref{thm2: bd8} from $r= 0$ to $R-1$ and then dividing them by $R$ yields\begin{align}
	\frac{1}{R}\sum_{r=0}^{R - 1}\E[G^r] 
	\leq & \frac{1}{R}\sum_{r=0}^{R - 1}\bigg(\frac{D_3(\wt P^r- \wt P^{r+1})}{R} + \frac{D_4^r\sigma^2}{S}\bigg) \leq \frac{D_3(\wt P^0-\ul f)}{R} + \frac{D_4\sigma^2}{\sqrt{R}}, \label{thm2: bd9}
\end{align}
where  $D_3 \triangleq \max\limits_{r} D_3^{r}, D_4 \triangleq \max\limits_{r} D_4^{r}$; \eqref{thm2: bd9} follows by Assumption 1 and the fact that $S = \sqrt{R}$.

\bibliography{refs20}

\end{document}